\documentclass[10pt,a4paper]{article}
\ifdefined\pdfminorversion\pdfminorversion=7\fi
\usepackage[a4paper,left=25mm,right=25mm,top=18mm,bottom=23mm,headheight=27pt,headsep=5pt,footskip=23pt]{geometry}
\usepackage[T1]{fontenc}
\usepackage{lmodern}
\usepackage{tgpagella}
\usepackage{amsmath,amssymb,amsthm,mathtools,graphicx,booktabs,array,microtype}
\usepackage{subcaption,multirow,longtable}
\usepackage{xurl}
\usepackage{titlesec}
\usepackage{fancyhdr,flafter,placeins,needspace}
\usepackage{xcolor}
\usepackage[authoryear,round]{natbib}
\definecolor{LinksBlue}{RGB}{34,63,142}
\usepackage[colorlinks=true,citecolor=LinksBlue,linkcolor=LinksBlue,urlcolor=LinksBlue]{hyperref}
\usepackage[capitalize,noabbrev]{cleveref}

\AtBeginDocument{%
  \setlength{\abovedisplayskip}{5pt plus 2pt minus 2pt}%
  \setlength{\belowdisplayskip}{5pt plus 2pt minus 2pt}%
  \setlength{\abovedisplayshortskip}{3pt plus 1pt minus 1pt}%
  \setlength{\belowdisplayshortskip}{4pt plus 1pt minus 1pt}%
}

\DeclareCaptionFont{talacaption}{\linespread{1}\fontsize{9}{10}\selectfont}
\titleformat{\section}{\normalfont\bfseries\fontsize{12}{13}\selectfont}{\thesection}{.65em}{}
\titleformat{\subsection}{\normalfont\bfseries\fontsize{10}{11}\selectfont}{\thesubsection}{.65em}{}
\titleformat{\subsubsection}{\normalfont\bfseries\fontsize{10}{11}\selectfont}{\thesubsubsection}{.65em}{}
\titleformat{\paragraph}[runin]{\normalfont\normalsize\bfseries}{}{0pt}{}
\titlespacing*{\section}{0pt}{11pt plus 2pt minus 2pt}{5pt plus 1pt minus 1pt}
\titlespacing*{\subsection}{0pt}{8pt plus 2pt minus 2pt}{4pt plus 1pt minus 1pt}
\titlespacing*{\subsubsection}{0pt}{6pt plus 1pt minus 1pt}{3pt}
\titlespacing*{\paragraph}{0pt}{5pt plus 1pt minus 1pt}{.45em}

\renewcommand{\headrulewidth}{0pt}
\renewcommand{\footrulewidth}{0pt}
\fancypagestyle{tala-first}{%
  \renewcommand{\headrulewidth}{.35pt}%
  \fancyhead[L]{\normalfont\linespread{1}\fontsize{9}{10}\selectfont\bfseries TALA}%
  \fancyhead[R]{\linespread{1}\fontsize{9}{10}\selectfont 2026-11-29}%
}

\makeatletter
\renewcommand{\@maketitle}{%
  \vspace*{1pt}%
  {\centering
    \linespread{1}\fontsize{14.4}{16}\selectfont\bfseries\@title\par
    \vspace{6pt}%
    \normalsize\bfseries\@author\par
  }%
  \vspace{9pt}%
}
\makeatother
\renewenvironment{abstract}{%
  {\centering\linespread{1}\fontsize{12}{13}\selectfont\bfseries Abstract\par}%
  \vspace{4pt}%
  \begin{list}{}{%
    \setlength{\leftmargin}{40pt}%
    \setlength{\rightmargin}{40pt}%
    \setlength{\topsep}{0pt}%
    \setlength{\partopsep}{0pt}%
  }\item\relax
}{\end{list}}

\theoremstyle{plain}

\theoremstyle{definition}

\theoremstyle{remark}

\definecolor{talaBaseInk}{HTML}{9A5E35}
\definecolor{talaStageInk}{HTML}{376B93}
\newcommand{\TALAb}{\textcolor{talaBaseInk}{\mathbf b}}
\newcommand{\TALAs}{\textcolor{talaStageInk}{\mathbf s_{\ell}^{(r)}}}

\input{tala_layout_style}

\TALAPreprinttrue
\title{Breaking the KV Cache Bottleneck in Looped Transformers with Shared Token Latents}
\hypersetup{pdftitle={Breaking the KV Cache Bottleneck in Looped Transformers with Shared Token Latents}}
\author{\mbox{Dehong Zheng\textsuperscript{1,2}},
\mbox{Tong Zhu\textsuperscript{2*}},
\mbox{Yunkai Chai\textsuperscript{2,3}},
\mbox{Xiaoyang Chen\textsuperscript{2,4}},
\mbox{Zefeng He\textsuperscript{2,3}},
\mbox{Zhe Xu\textsuperscript{2}},
\mbox{Xuyang Hu\textsuperscript{2}},
\mbox{Xinyu Zhang\textsuperscript{2,1}},
\mbox{Zongliang Wu\textsuperscript{2}},
\mbox{Zhongzhi Zhang\textsuperscript{1}},
\mbox{Yafu Li\textsuperscript{2,5}},
\mbox{Yu Cheng\textsuperscript{6*}}\par\smallskip
{\normalfont\small
\textsuperscript{1}Fudan University,
\textsuperscript{2}Shanghai AI Laboratory,
\textsuperscript{3}Shanghai Jiao Tong University,
\textsuperscript{4}University of Science and Technology of China,
\textsuperscript{5}The Chinese University of Hong Kong,
\textsuperscript{6}Nanyang Technological University\par\smallskip
\href{mailto:23307130129@m.fudan.edu.cn}{\texttt{23307130129@m.fudan.edu.cn}},
\href{mailto:zhutong@pjlab.org.cn}{\texttt{zhutong@pjlab.org.cn}},
\href{mailto:yu.cheng@ntu.edu.sg}{\texttt{yu.cheng@ntu.edu.sg}}\par}
}

\date{}
\begin{document}
\maketitle
\thispagestyle{tala-first}
\begingroup
\renewcommand{\thefootnote}{}
\footnotetext[0]{This work was conducted during Dehong Zheng's internship at Shanghai AI Laboratory.\par
\textsuperscript{*}\,Corresponding authors.\par
Code: \url{https://github.com/ddshif/TALA}}
\endgroup

\begin{abstract}
Looped Transformers offer a promising way to scale computational depth at a fixed parameter count, yet their key--value (KV) cache grows linearly with recursion depth under standard attention. Since different recursions process the same sequence, their KV representations may encode much of the same information, alongside information specific to each recursion. To investigate this possibility, we learn memory-sharing policies during pretraining and find that under simulated storage pressure, early KV representations are broadly reused while intermediate representations remain useful. This observation motivates separating KV representations into a shared base and compact stage-specific latents. Interestingly, we further find that extending base sharing across physical layers improves model quality while reducing KV storage. Based on these findings, we propose \textbf{Token-Anchored Latent Attention (TALA)}, which derives a single base latent from each token's initial representation, shares it across all layers and recursions, and complements it with compact stage latents. Each physical layer decodes the base and its current stage into keys and values, adding their contributions before attention. \textbf{\boldmath Compared with vanilla looped Transformers in equal-token pretraining, TALA achieves approximately $12\times$ KV cache compression and uses 28.4\% fewer training FLOPs, while attaining lower validation NLL and higher mean scores on commonsense and RULER-NIAH benchmarks.} Our IsoFLOPs scaling experiments demonstrate consistent superiority over all three baselines across compute budgets, with no predicted crossover in the fitted compute-optimal frontiers. Training trajectories further show that TALA achieves $1.92\times$ the training compute efficiency of vanilla looped Transformers at a matched training NLL. Its advantages grow under aggressive KV compression and extend to conventional Transformers. Our mechanistic analyses further suggest that shared token latents absorb reusable token-specific matching functions, reducing representational redundancy and freeing stage latent capacity for contextual information.
\end{abstract}

\newcommand{\TALAArchitectureText}{Token-anchored latent attention. A token-derived base is shared across layers and recursions (left). Each layer execution adds its decoded base and stage keys and values before attention (right). Projection weights are shared across recursions. The figure shows the content pathway.}
\ifTALAPreprint
  \par
  \noindent\begin{minipage}{\textwidth}
    \centering
    \includegraphics[width=.74\linewidth]{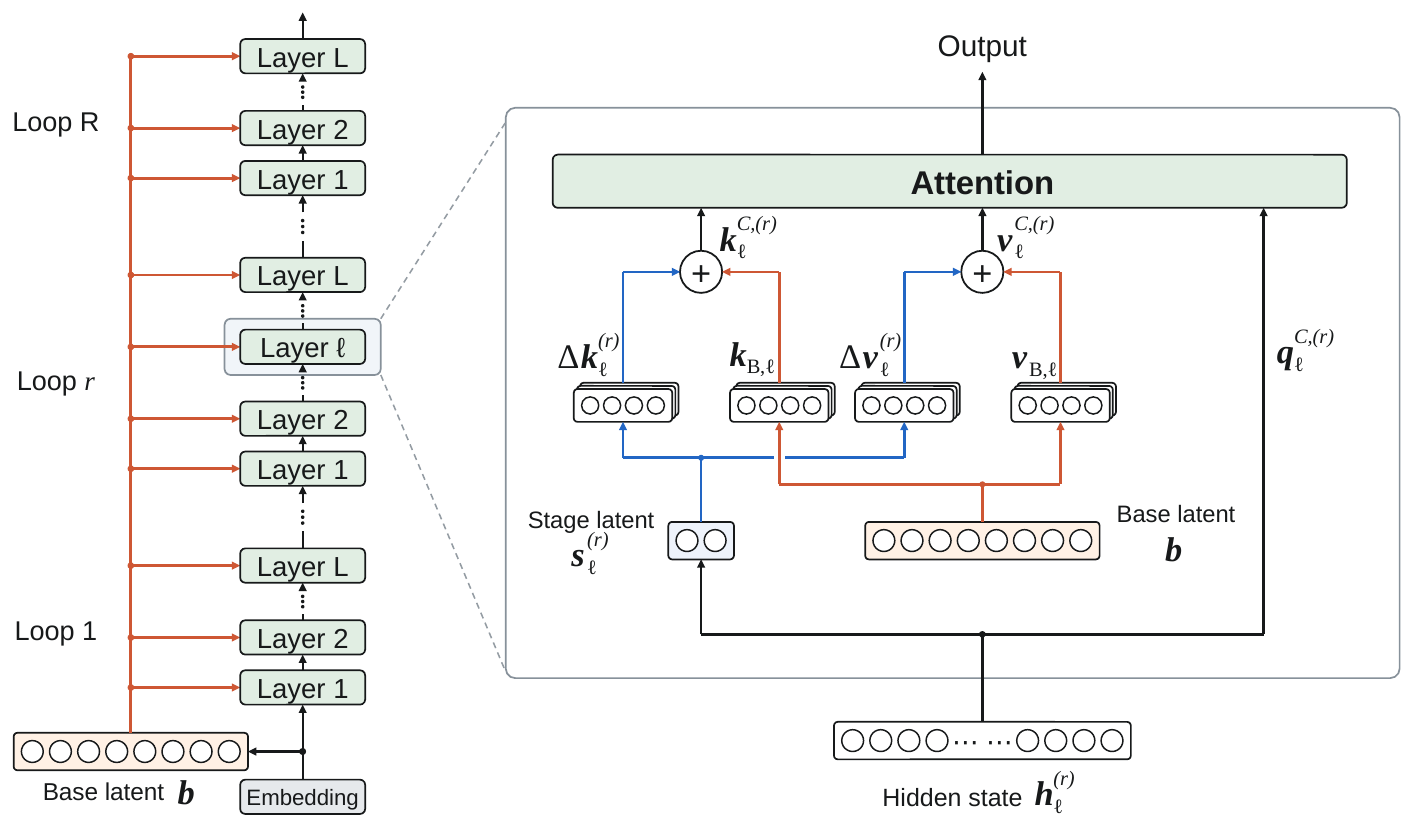}
    \captionsetup{hypcap=false,width=.90\linewidth}
    \captionof{figure}{\TALAArchitectureText}
    \label{fig:tala-architecture}
  \end{minipage}
  \par
\else
  \begin{figure}[t]
    \centering
    \includegraphics[width=\linewidth]{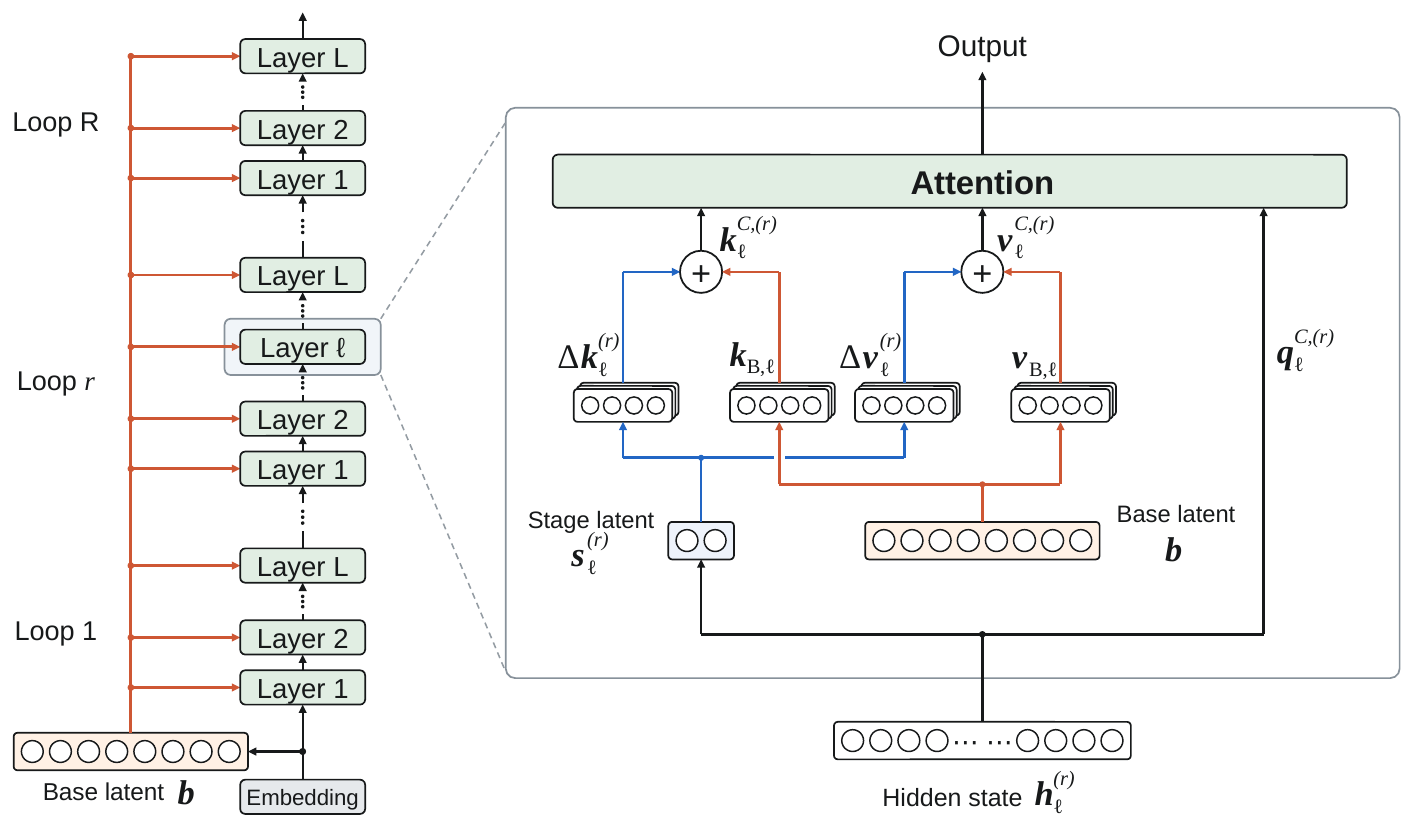}
    \caption{\TALAArchitectureText}
    \label{fig:tala-architecture}
  \end{figure}
\fi

\clearpage
\section{Introduction}\label{sec:introduction}

Looped Transformers (LT) offer a promising way to improve language model quality through additional computation at a fixed parameter count~\citep{universal_transformer_2019,huginn_2025,ouro_2025}. By repeatedly applying a shared stack of layers, they increase the effective depth of the network without increasing the memory required to store model weights. However, the hidden states evolve across recursions, and standard attention maintains a separate key--value (KV) cache at each layer in each recursion. For example, a 12-layer Transformer with four recursions stores the weights of only 12 layers but maintains KV caches for 48 layer executions. Consequently, KV storage grows linearly with recursion depth, diminishing the memory advantage of looped architectures, particularly at long context lengths and large decoding batch sizes.

Recent studies have explored various approaches to reducing KV storage in looped Transformers. Methods operating on pretrained models exploit redundancy across recursions through KV quantization, sparse updates, and cache reuse~\citep{flashloop_2026,loopkv_2026,loopvq_2026}. Other approaches employ additional training to adapt models to shared KV caches~\citep{melt_2026,benice_2026}. Cross-recursion memory sharing has also been incorporated directly into model architectures. PLT~\citep{plt_2025} and LPT~\citep{lpt_2026} share the first-recursion KV history across subsequent recursions and supplement it with local attention. LPT demonstrates that such memory sharing can improve language modeling quality while substantially reducing KV storage. A similar global--local memory organization has recently been adopted in GLM-5.4~\citep{zai2026glm54} through continued training. The effectiveness of cross-recursion memory sharing suggests that different recursions can benefit from common information about historical tokens. Meanwhile, the global--local design restricts direct access to KV representations produced by later recursions to a local window. These observations motivate us to examine the organization of KV memory at a finer granularity: \emph{what information in a token's KV representations should be shared across computation stages, and what should be stored separately?}

To investigate this question, we train looped Transformers to learn how to reuse KV representations across recursions. We allow each recursion to combine causally available KV histories and introduce a differentiable storage penalty to encourage memory reuse during pretraining. Under simulated storage pressure, later recursions increasingly rely on the first-recursion KV history while continuing to draw substantial contributions from intermediate recursions (Figure~\ref{fig:softbank_weights}). This observation suggests that early representations contain features useful across multiple recursions, while subsequent computation provides additional information that complements these features. It motivates a simple design in which each physical layer maintains a relatively wide base latent shared across recursions, together with compact stage latents derived from the hidden states at individual recursions. Both components adopt the latent KV parameterization of multi-head latent attention (MLA; \citealp{deepseekv2_2024}).

The utility of a shared base across recursions further motivates us to explore sharing across physical layers. Interestingly, progressively extending the sharing scope of the base consistently improves validation quality while reducing KV storage (Figure~\ref{fig:tala-sharing-span}). The best-performing configuration shares a single base across the entire network, constructed directly from the initial token representation before any contextual attention. Although the contextual representation of a token evolves throughout the network, its identity remains unchanged. Providing token-dependent features through a shared representation may therefore reduce the need to encode them repeatedly at different computation stages.

Based on these observations, we propose \textbf{Token-Anchored Latent Attention (TALA)}, an attention architecture that shares a single token-derived base latent across all physical layers and recursions (Figure~\ref{fig:tala-architecture}). At each recursion, every physical layer constructs a compact stage latent from its current hidden state. The base and stage latents are decoded into keys and values using layer-specific projections, and their contributions are added before attention. All projection parameters are shared across recursions, preserving the parameter-sharing principle of looped Transformers. Different physical layers can extract distinct features from the same shared base, while evolving queries and stage latents enable recursion-specific attention. In particular, stage keys refine the retrieval of historical tokens, and stage values complement the content provided by the base. By amortizing the storage of a relatively wide base across all layers and recursions, TALA substantially reduces the KV capacity required at each layer execution.

Our IsoFLOPs scaling experiments show that TALA achieves the lowest validation NLL at all four compute budgets, with its fitted compute-optimal frontier consistently outperforming all three baselines and exhibiting no predicted crossover as training compute increases. At a validation NLL of 2.65, the fitted frontiers estimate training compute savings of 23.7\%, 32.2\%, and 57.2\% relative to LT-MLA, LPT-MLA, and vanilla-LT, respectively. For approximately 450M-parameter models with four recursions trained on 20B tokens, TALA reduces validation NLL by up to 0.041, improves mean scores on nine commonsense benchmarks by up to 2.01 points and on RULER-NIAH by up to 17.11 points relative to the baselines. \textbf{\boldmath Compared with vanilla-LT, TALA achieves approximately $12\times$ KV cache compression and uses 28.4\% fewer training FLOPs, while attaining lower validation NLL and higher mean scores on commonsense and RULER-NIAH benchmarks.} Its advantages persist across recursion depths and become more pronounced under aggressive KV compression, with TALA at $36\times$ compression achieving lower NLL than LT-MLA at $6\times$ compression. We further apply TALA to conventional Transformers, where it outperforms eight standard attention baselines, reducing validation NLL by up to 0.045 and improving mean scores on nine commonsense benchmarks by up to 2.06 points and on RULER-NIAH by up to 49.91 points. Our attention matching analysis suggests that the shared base takes on reusable token-specific matching functions, reducing redundant encoding across computation stages and freeing representational capacity in compact stage latents for contextual information.

\section{Preliminaries}\label{sec:preliminaries}
\subsection{Looped Transformers}\label{sec:prelim_lt}
A looped Transformer repeatedly applies a shared stack of $L$ physical layers for $R$ recursions, achieving an effective depth of $LR$ while retaining the parameters of an $L$-layer network. Let $\mathcal F_\theta$ denote one traversal of the shared stack, including its recurrent-state update, and $\mathbf H^{(0)}$ the initial token representations. The recurrent computation is
\begin{equation}\label{eq:prelim_loop}
\mathbf H^{(r)}=\mathcal F_\theta(\mathbf H^{(r-1)}),\qquad r=1,\ldots,R.
\end{equation}
We index physical layers by $\ell$, recursions by $r$, and query and source positions by $i$ and $j$, respectively. Let $\mathbf h_{\ell,j}^{(r)}$ denote the input to attention at layer $\ell$ in recursion $r$. Although the attention parameters are shared across recursions, the hidden states evolve with computation, and each recursion constructs its own keys and values. Under causal self-attention~\citep{vaswani2017attention}, a query at position $i$ accesses
\begin{equation}\label{eq:prelim_lt_bank}
\mathcal C_\ell^{(r)}(i)=\{(\mathbf k_{\ell,j}^{(r)},\mathbf v_{\ell,j}^{(r)}):j\leq i\}.
\end{equation}
During autoregressive decoding, a conventional looped Transformer therefore maintains $R$ separate KV histories per physical layer, with total KV storage growing linearly with recursion depth.

\subsection{Multi-head Latent Attention}\label{sec:prelim_mla}
Multi-head latent attention (MLA) compresses the keys and values of all attention heads into a shared low-dimensional latent to reduce the KV cache~\citep{deepseekv2_2024}. At a given layer execution, we omit the layer and recursion indices and denote the attention input at position $j$ by $\mathbf h_j\in\mathbb R^d$. MLA parameterizes the content keys and values as
\begin{equation}\label{eq:prelim_mla}
\mathbf c_j^{KV}=W^{DKV}\mathbf h_j,\quad \mathbf k_j^C=W^{UK}\mathbf c_j^{KV},\quad \mathbf v_j^C=W^{UV}\mathbf c_j^{KV}.
\end{equation}
Here, $\mathbf c_j^{KV}\in\mathbb R^{d_c}$ is the shared latent, $W^{DKV}\in\mathbb R^{d_c\times d}$ is the down-projection, and $W^{UK},W^{UV}\in\mathbb R^{n_hd_h\times d_c}$ are the key and value up-projections, where $n_h$ is the number of attention heads and $d_h$ is the content dimension per head. With $d_c\ll n_hd_h$, caching the latent reduces content KV storage per token from $2n_hd_h$ to $d_c$ elements. We omit the decoupled positional key for clarity. In a conventional looped Transformer, each layer execution still produces and caches its own latent.

\section{Shared Memory Across Recurrence and Depth}\label{sec:method}\label{sec:shared_base_stage}
\subsection{Discovering Causal Memory Reuse}\label{sec:method_discovery}
To understand how KV memory should be shared across recursions, we jointly learn the model and its memory-sharing policy during pretraining. Under standard token-parallel training, each recursion processes all token positions in parallel, layer by layer, before the next recursion begins. At a given physical layer, recursion $r$ can therefore access only KV histories produced at recursions $s\leq r$, since the corresponding states of later recursions have not yet been computed (Figure~\ref{fig:softbank_weights}a).

\TALAFloat{figure}{!t}{!ht}
\centering
\includegraphics[width=\linewidth]{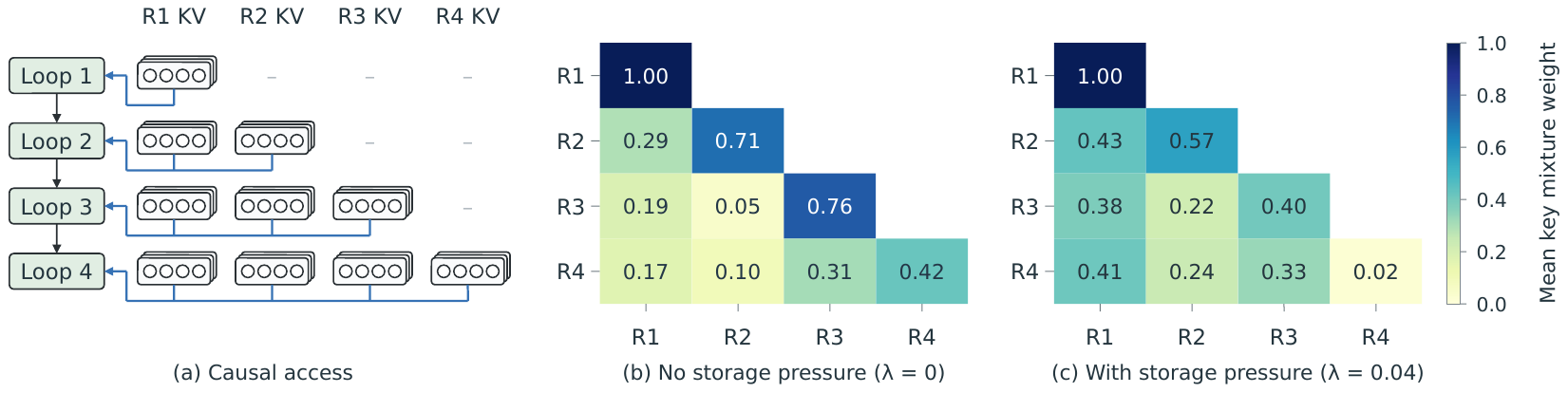}
\caption{Causal memory access and learned reuse. (a) Recursion $r$ can read KV banks from recursions $s\leq r$. (b,c) Mean key-mixture weights with $\lambda=0$ and $0.04$, respectively. Rows denote readers and columns denote writers. Both models retain all four banks.}\label{fig:softbank_weights}\label{fig:softbank-donor}
\end{figure}

\paragraph{Learning cross-recursion memory reuse.}
We train a looped Transformer from scratch, allowing each recursion to construct its own KV history, referred to as a memory bank. At a fixed physical layer and attention head, let $\mathbf k_j^{(s)}$ and $\mathbf v_j^{(s)}$ denote the key and value at source position $j$ produced by recursion $s$. Each recursion $r$ learns to combine the available banks:
\begin{equation}\label{eq:softbank_read}
\widetilde{\mathbf k}_j^{(r)}=\sum_{s=1}^{r}\alpha_{r,s}^{K}\mathbf k_j^{(s)},\qquad
\widetilde{\mathbf v}_j^{(r)}=\sum_{s=1}^{r}\alpha_{r,s}^{V}\mathbf v_j^{(s)}.
\end{equation}
The mixture weights $\alpha_{r,s}^{K}$ and $\alpha_{r,s}^{V}$ represent the contributions of bank $s$ to recursion $r$. They are parameterized by learnable logits and normalized over causally available banks:
\begin{equation}\label{eq:bank-mixture}
\alpha_{r,s}^{X}=\frac{\exp(\theta_{r,s}^{X}/\tau)}{\sum_{t=1}^{r}\exp(\theta_{r,t}^{X}/\tau)},\qquad X\in\{K,V\}.
\end{equation}
We fix the softmax temperature to $\tau=1$ and initialize all logits to zero, yielding uniform weights over available banks. The weights are learned independently for each layer and head and shared across token positions. The mixed keys and values are used in a single causal attention operation.

\paragraph{Simulating memory pressure.}
Without a storage constraint, each recursion can freely retain and use its own KV bank, and the training objective provides no explicit incentive for memory reuse. To investigate which histories are preferentially retained under limited storage, we introduce a differentiable storage penalty. The underlying principle is that a bank shared by multiple readers needs to be stored only once.

Since mixture weights alone may be confounded by the magnitude of their associated representations, we first compute scale-adjusted usage coefficients:
\begin{equation}\label{eq:bank-rho}
\rho_{r,s}^{X}=\frac{\alpha_{r,s}^{X}\operatorname{sg}(m_s^{X}+\epsilon)}{\sum_{t=1}^{r}\alpha_{r,t}^{X}\operatorname{sg}(m_t^{X}+\epsilon)+\epsilon},
\end{equation}
where $m_s^{X}$ is an exponential moving average of the RMS magnitude of bank $s$, $\operatorname{sg}(\cdot)$ denotes the stop-gradient operator, and $\epsilon$ is a small constant for numerical stability. These coefficients account for both mixture weights and representation magnitudes when estimating bank usage.

We then approximate the collective usage of bank $s$ across all eligible readers through a soft union:
\begin{equation}\label{eq:bank-union}
u_s=1-\prod_{r=s}^{R}\bigl[(1-\rho_{r,s}^{K})(1-\rho_{r,s}^{V})\bigr].
\end{equation}
This expression is a differentiable relaxation of the logical OR. In the binary case, a bank is needed if at least one reader uses its keys or values. Each factor $(1-\rho_{r,s}^{K})(1-\rho_{r,s}^{V})$ approaches one when reader $r$ makes little use of either component. The product across eligible readers captures collective non-usage, so subtracting it from one gives a soft retention cost bounded by one, regardless of the number of readers.

Restoring the layer and head indices, we sum the retention costs over banks and average across layers and heads:
\begin{equation}\label{eq:bank-cost}
\widehat B_{\mathrm{soft}}=\frac{1}{Ln_h}\sum_{\ell=1}^{L}\sum_{a=1}^{n_h}\sum_{s=1}^{R}u_{\ell,a,s}.
\end{equation}
Here, $\widehat B_{\mathrm{soft}}$ measures the effective number of retained banks through a continuous proxy. We incorporate the normalized storage penalty into the training objective:
\begin{equation}\label{eq:bank-objective}
\mathcal L=\mathcal L_{\mathrm{NTP}}+\lambda\frac{\widehat B_{\mathrm{soft}}-1}{R-1},
\end{equation}
where $\mathcal L_{\mathrm{NTP}}$ is the next-token prediction loss and $\lambda$ controls the storage pressure. All KV banks remain physically stored during training, allowing the model to explore memory-sharing patterns without imposing a predetermined retention policy.

\paragraph{Observations and implications.}
We train two approximately 150M-parameter, four-recursion models on 10B tokens, with $\lambda=0$ and $\lambda=0.04$, respectively. Without storage pressure, each recursion assigns the largest mean key-mixture weight to its own bank (Figure~\ref{fig:softbank_weights}b). Introducing the storage penalty reduces the effective bank count from 3.412 to 2.512 while maintaining comparable validation quality. The first-recursion bank receives substantially higher mixture weights, while intermediate-recursion banks retain considerable mixture weights (Figure~\ref{fig:softbank_weights}c). Appendix~\ref{app:soft_budget} reports the validation metrics.

We interpret these observations through the interplay between \emph{reuse opportunity} and \emph{representation utility}: earlier histories can serve more readers, while later histories may contain valuable information formed through additional computation. This motivates a finer-grained approach to memory reuse, in which a common component of historical KV representations is shared across recursions and compact stage-specific components preserve additional information.

\subsection{Token-Anchored Latent Attention}\label{sec:method_architecture}

\paragraph{Sharing across recursions.}
The observations in Section~\ref{sec:method_discovery} motivate an asymmetric allocation of KV capacity across recursions. We construct a relatively wide latent in the first recursion as a shared base, complemented by a compact stage latent at each recursion to capture information specific to its computation. Following the MLA formulation, we fix a source token and omit its position index. At physical layer $\ell$, the base and stage latents are
\begin{equation}\label{eq:initial_stage_write}
\mathbf b_\ell=W_{\ell,B}^{DKV}\mathbf h_\ell^{(1)},\qquad
\mathbf s_\ell^{(r)}=W_{\ell,S}^{DKV}\mathbf h_\ell^{(r)},\quad 1\leq r\leq R.
\end{equation}
Here, $\mathbf b_\ell\in\mathbb R^{d_B}$ and $\mathbf s_\ell^{(r)}\in\mathbb R^{d_S}$ are the base and stage latents, respectively, with $d_S<d_B$ in our design. The subscripts $B$ and $S$ distinguish their projection matrices. The content keys and values are given by
\begin{equation}\label{eq:initial_stage_read}
\mathbf k_\ell^{C,(r)}=W_{\ell,B}^{UK}\mathbf b_\ell+W_{\ell,S}^{UK}\mathbf s_\ell^{(r)},\qquad
\mathbf v_\ell^{C,(r)}=W_{\ell,B}^{UV}\mathbf b_\ell+W_{\ell,S}^{UV}\mathbf s_\ell^{(r)}.
\end{equation}
At a fixed physical layer, the decoded base keys and values remain identical across recursions, while each recursion adds its own stage contribution at every historical position. This allows the model to reuse a common representation while preserving compact, recursion-specific information throughout the full context.

\paragraph{Sharing across physical layers.}
Since the base is intended to encode information that remains useful across successive computations, we extend its sharing scope to multiple physical layers. To accommodate their different representational requirements, each layer retains its own key and value up-projections. We partition the $L$ physical layers into $L/G$ groups of $G$ consecutive layers, where $G$ divides $L$. Let $\mathcal G_g$ denote the $g$-th group and $\ell_g$ its first physical layer. A single base is constructed for each group during the first recursion:
\begin{equation}\label{eq:group_base}
\begin{aligned}
\mathbf b_g&=W_{g,B}^{DKV}\mathbf h_{\ell_g}^{(1)},\\
\mathbf k_{B,\ell}&=W_{\ell,B}^{UK}\mathbf b_g,\qquad
\mathbf v_{B,\ell}=W_{\ell,B}^{UV}\mathbf b_g,\quad \ell\in\mathcal G_g.
\end{aligned}
\end{equation}
Each base latent is thus shared across $GR$ layer executions, with layer-specific decoding projections that are reused across recursions. The stage remains specific to each layer execution, as in Equations~\eqref{eq:initial_stage_write}--\eqref{eq:initial_stage_read}.

To examine how broadly a base can be shared, we vary the sharing span $G\in\{1,2,4,6,12\}$ in a twelve-layer, four-recursion model, keeping the base and stage widths and training-token budget fixed. Interestingly, increasing the sharing span consistently improves validation quality across the tested configurations while reducing persistent KV storage. Extending the sharing span from one physical layer to the entire stack reduces validation NLL from 2.982 to 2.941 and persistent KV storage by approximately 44\% (Figure~\ref{fig:tala-sharing-span}). These results motivate full-stack sharing ($G=L$), where a single base is constructed directly from the initial token representation, before any contextual attention. Appendix~\ref{sec:sharing-span-details} reports the detailed resource and quality measurements.

\TALAFloat{figure}{!ht}{!ht}
\centering\includegraphics[width=.97\linewidth]{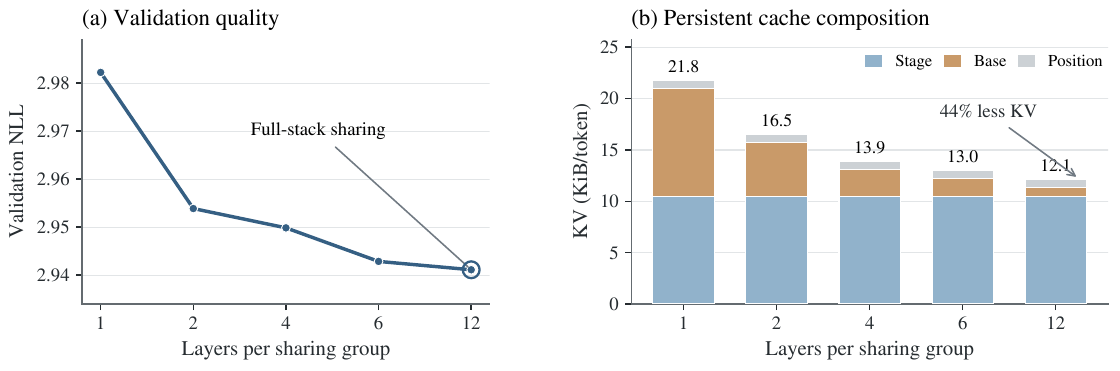}
\caption{Effect of base-sharing span ($L=12$, $R=4$, base/stage widths 448/112). (a) Validation NLL. (b) Whole-model persistent BF16 KV storage per token, decomposed into base, stage, and positional states.}\label{fig:tala-sharing-span}
\end{figure}

\paragraph{Token-anchored latent attention.}
With full-stack sharing, the base provides a stable, token-specific representation throughout the network, complemented by a compact stage derived from the current hidden state at each layer execution. We refer to this architecture as \emph{Token-Anchored Latent Attention} (TALA). Let $\mathbf h^{(0)}$ denote the initial representation of the source token. The base and stage latents are
\begin{equation}\label{eq:base_stage_write}
\TALAb=W_B^{DKV}\mathbf h^{(0)},\qquad
\TALAs=W_{\ell,S}^{DKV}\mathbf h_\ell^{(r)},\quad 1\leq r\leq R,
\end{equation}
and the corresponding content keys and values are
\begin{equation}\label{eq:base_stage_decode}
\mathbf k_\ell^{C,(r)}=
\underbrace{W_{\ell,B}^{UK}\TALAb}_{\text{shared base}}
+\underbrace{W_{\ell,S}^{UK}\TALAs}_{\text{current stage}},\qquad
\mathbf v_\ell^{C,(r)}=
\underbrace{W_{\ell,B}^{UV}\TALAb}_{\text{shared base}}
+\underbrace{W_{\ell,S}^{UV}\TALAs}_{\text{current stage}}.
\end{equation}
The base is derived from the existing token representation without introducing an additional learned token table. The base down-projection is global, while each physical layer has its own base up-projections and stage down/up-projections. All projection parameters are shared across recursions. Consequently, different physical layers can extract distinct features from the same base, while the decoded base keys and values at a fixed physical layer remain unchanged across recursions. Equations~\eqref{eq:base_stage_write}--\eqref{eq:base_stage_decode} define a shared KV representation that can be combined with different memory-retention policies.

\paragraph{Full-history memory.}
Our default architecture retains the stage latents of every layer execution. Each reader combines the shared base with its own stage at each historical position before a single causal softmax (Figure~\ref{fig:tala-architecture}). Previous stages are not accumulated. Unless otherwise specified, we use $d_B=4d_S$. Let $T$ be the context length, $\eta$ the number of bytes per stored scalar, and $d_p$ the width of the positional state shared across recursions at each physical layer. The persistent KV storage under grouped and full-stack sharing is
\begin{equation}\label{eq:shared_cache_cost}
M_G(T)=\eta T\left(\frac{L}{G}d_B+LRd_S+Ld_p\right),\qquad
M_{\mathrm{all}}(T)=\eta T\left(d_B+LRd_S+Ld_p\right).
\end{equation}
Full-stack sharing amortizes the base storage over all $LR$ layer executions, permitting a relatively wide base at a small amortized cache cost. The stage storage remains proportional to the number of layer executions. These expressions account for logical persistent cache storage, excluding reconstructed content tensors and temporary computation buffers.

\paragraph{Global--local memory.}
The additive KV representation also permits more aggressive memory retention. We retain the full-history stage latents of the first recursion and restrict subsequent recursions to their own local windows of decoded keys and values. Each later recursion jointly attends to the first-recursion global history and its current local window through a single softmax. Recent token positions may appear in both sets as distinct attention candidates. This variant, TALA+SWA, follows the memory-access pattern of LPT-MLA+SWA while using token-anchored KV representations. In our main comparison, we use a 64-token local window. Appendix~\ref{app:memory_details} provides the corresponding cache accounting.

\subsection{Reader-Specific Retrieval and Contextual Refinement}\label{sec:method_mechanism}

Although the base latent is shared across the entire network, different physical layers can extract distinct features through their own decoding projections, and queries evolve across recursions. The stage further refines both the retrieval of historical tokens and the content recovered from them. We illustrate these two roles in the full-history architecture.

\paragraph{Stage keys refine retrieval.}
Consider a fixed physical layer, recursion, attention head, and query at position $i$. Let $p_B(j)$ denote the attention distribution obtained using only the base keys and positional scores. With the query, positional scores, and causal mask fixed, adding the stage key introduces a score correction
\[
\delta_j=\frac{(\mathbf q_i^C)^\top\Delta\mathbf k_j}{\sqrt{d_a}},
\]
where $\Delta\mathbf k_j$ is the decoded stage key for the corresponding head and $d_a$ is the attention-score scaling dimension. The resulting attention distribution is
\begin{equation}\label{eq:stage_exponential_tilt}
p_F(j)=\frac{p_B(j)\exp(\delta_j)}{\mathbb E_{t\sim p_B}[\exp(\delta_t)]}.
\end{equation}
The stage key thus performs a query-dependent exponential reweighting of the base retrieval distribution. By modifying the relative attention scores of historical tokens, each stage can adapt retrieval to its current computation while reusing the matching features provided by the base.

\paragraph{Stage values refine retrieved content.}
Let $A_B$ and $A_F$ denote the base-only and full attention weight matrices under the same queries, positional scores, and causal mask. Given the decoded base values $V_B$ and stage values $\Delta V$, the corresponding attention outputs before the output projection are $O_B=A_BV_B$ and $O_F=A_F(V_B+\Delta V)$. Their difference can be decomposed as
\begin{equation}\label{eq:stage_value_identity}
O_F-O_B=\underbrace{(A_F-A_B)V_B}_{\text{retrieval refinement}}+\underbrace{A_F\Delta V}_{\text{content refinement}}.
\end{equation}
The first term captures how stage keys change the retrieval of base values, while the second captures the additional content supplied by stage values under the resulting attention distribution.

These properties suggest a functional specialization in which the shared base supports reusable token-specific matching and content, while compact stages adapt retrieval and retrieved content to successive computations. Section~\ref{sec:exp_mechanism} examines whether this specialization emerges in trained models.

\section{Experiments}\label{sec:experiments}
\subsection{Experimental Setup}\label{sec:exp_setup}
\paragraph{Training settings.} We pretrain decoder-only Transformers on FineWeb-Edu~\citep{penedo2024fineweb} with a context length of 4,096, using AdamW~\citep{loshchilov2019adamw} and a warmup--stable--decay (WSD) learning rate schedule~\citep{hu2024minicpm}. Unless otherwise specified, looped models use four recursions, adopt update normalization~\citep{lpt_2026}, and compute the next-token prediction loss only at the final recursion.

For compute-optimal scaling, we vary physical depth and training duration across four compute budgets: $3\times10^{18}$, $10^{19}$, $3\times10^{19}$, and $10^{20}$ FLOPs. In a separate fixed-depth comparison, we train models with approximately 450M parameters and four recursions on 20B tokens. Experiments on sharing span, recursion depth, and KV compression use approximately 150M-parameter models trained on 3B tokens. We also train non-loop Transformers on 10B tokens. Detailed training configurations are provided in Appendices~\ref{app:exp_protocol} and~\ref{app:exp_nonloop_details}.

\paragraph{Baselines.} Our primary scaling comparison consists of four full-history architectures: vanilla-LT, which uses standard multi-head attention (MHA) and maintains independent KV histories at every layer execution; LT-MLA, which independently caches an MLA latent at every layer execution; LPT-MLA, which shares the first-recursion latent of each physical layer across subsequent recursions, following the early-memory sharing principle of LPT~\citep{lpt_2026}; and TALA, which shares a global base complemented by compact stage-specific latents. The three latent attention architectures target approximately 12-fold compression relative to dense content KV at each matched physical depth. We additionally compare LT-MLA (wide), which uses four times the cache capacity of LT-MLA at the same physical depth.

To examine compatibility with global--local memory sharing, we include LPT-MLA+SWA, which combines the first-recursion MLA history with a 64-token sliding window of full keys and values at each subsequent recursion through a joint softmax. This baseline adapts the global--local memory organization used in LPT and GLM-5.4~\citep{lpt_2026,zai2026glm54} to dense attention with MLA. TALA+SWA adopts the same memory-access policy with our token-anchored KV representations.

\paragraph{Evaluation.} We measure language modeling quality using validation negative log-likelihood (NLL). We evaluate zero-shot performance using the LM Evaluation Harness~\citep{gao_lm_eval_harness,biderman2024lessons} on nine language benchmarks: LAMBADA OpenAI and Standard~\citep{paperno2016lambada}, HellaSwag~\citep{zellers2019hellaswag}, PIQA~\citep{bisk2020piqa}, WinoGrande~\citep{sakaguchi2020winogrande}, ARC-Easy and ARC-Challenge~\citep{clark2018arc}, SciQ~\citep{welbl2017sciq}, and MMLU-continuation~\citep{hendrycks2021mmlu}. We further evaluate six needle-in-a-haystack (NIAH) tasks from RULER~\citep{hsieh2024ruler} at a context length of 4,096: three single-needle variants (S1, S2, S3), multi-key retrieval (MK1), multi-value retrieval (MV), and multi-query retrieval (MQ). We report the unweighted average within each benchmark group.

\paragraph{Resource accounting.} We report total training FLOPs and persistent BF16 KV storage. KV compression is measured relative to the corresponding dense-attention baseline at a context length of 4,096. Appendix~\ref{app:memory_details} provides detailed resource accounting.

\subsection{Main Results}\label{sec:exp_main}
\paragraph{Compute-optimal scaling.}
We investigate the compute efficiency of four full-history architectures: vanilla-LT, LT-MLA, LPT-MLA, and TALA. At each compute budget, we vary the model size and number of training tokens, using six model sizes at each of the first three budgets and five at the largest budget, yielding a total of 92 model--data allocations. The three latent attention architectures target approximately 12-fold KV compression relative to vanilla-LT at each matched physical depth.

Following the parametric scaling approach of \citet{hoffmann2022training} and recent work on recurrent-model scaling~\citep{he2026latentloop,wang2026smelt}, we fit a separate loss surface for each architecture:
\begin{equation}\label{eq:exp_scaling_fit}
\widehat{\mathcal L}(F,D)=E+A\left(\frac{F}{10^9}\right)^{-\alpha}+K\left(\frac{D}{10^{10}}\right)^{-\beta},\qquad C=FD.
\end{equation}
\newcommand{\TALAScalingFigureCaption}{Compute-optimal scaling. Solid and dashed curves denote fitted optima within and beyond the observed allocation hull, respectively. The inset compares estimated compute requirements at NLL 2.65, including the extrapolated vanilla-LT estimate (dashed arrow).}
\ifTALAPreprint\else
\begin{figure}[!htbp]
\centering
\includegraphics[width=.55\linewidth]{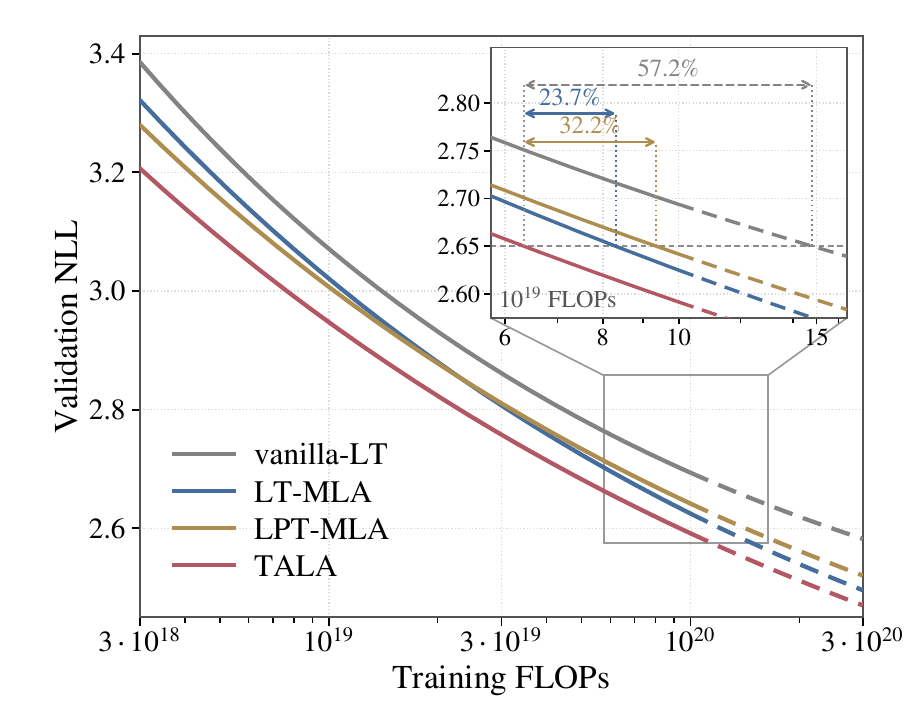}
\caption{\TALAScalingFigureCaption}\label{fig:scaling-frontier}
\end{figure}
\fi

Here, $F$ denotes training FLOPs per token, $D$ denotes the number of training tokens, and $C$ is the total training compute. We fit the five parameters using Huber loss on log-NLL residuals with a threshold of $10^{-3}$, employing bounded nonlinear least squares with 48 initializations. For a given compute budget $C$, we minimize the fitted loss subject to $FD=C$ to obtain the compute-optimal frontier. Appendix~\ref{app:exp_scaling_details} provides the detailed configurations and fitting procedure.

\paragraph{Scaling results.}
TALA achieves the lowest observed validation NLL at all four compute budgets. Its advantage holds across all 23 matched budget--depth configurations against each of the three baselines. As shown in Figure~\ref{fig:scaling-frontier}, the fitted compute-optimal frontier of TALA remains below those of all three baselines, with no predicted crossover as training compute increases. At a validation NLL of 2.65, the fitted frontiers estimate training-compute savings of 57.2\%, 23.7\%, and 32.2\% relative to vanilla-LT, LT-MLA, and LPT-MLA, respectively.

\ifTALAPreprint
\TALAFloat{table}{!t}{!htbp}
\centering
\caption{Checkpoints selected by validation NLL at each compute budget. $L$ is physical depth and Params is in millions. L-O/L-S denote LAMBADA OpenAI/Standard, HS HellaSwag, WG WinoGrande, and A-E/A-C ARC-Easy/Challenge. MMLU uses continuation scoring. Avg. and NIAH are the unweighted means (\%) of nine language benchmarks and six 4K RULER retrieval tasks. Bold marks the best unrounded value per budget. $\dagger$ At the two largest budgets, LPT-MLA scores come from separate evaluation runs with unverified sample equivalence to TALA.}\label{tab:scaling-selected}
\TALAtableStyle
\TALAScoredTable{.45pt}{l!{\hspace{\TALAPreprintValue{1.5pt}{0pt}}}rrr}{11}{4}{%
\toprule
\textbf{Model} & \multicolumn{1}{c}{\textbf{$L$}} & \multicolumn{1}{c}{\textbf{Params}} & \multicolumn{1}{c}{\mbox{\textbf{NLL}$\downarrow$}} & \TALAScoreHead{L-O} & \TALAScoreHead{L-S} & \TALAScoreHead{HS} & \TALAScoreHead{PIQA} & \TALAScoreHead{WG} & \TALAScoreHead{A-E} & \TALAScoreHead{A-C} & \TALAScoreHead{SciQ} & \TALAScoreHead{MMLU} & \TALAScoreHead[\uparrow]{Avg.} & \TALAScoreHead[\uparrow]{NIAH} \\
\midrule
\multicolumn{15}{l}{\textit{$3\times10^{18}$ FLOPs}} \\
vanilla-LT & 8 & 60 & 3.394 & 17.48 & 14.21 & 27.77 & 58.98 & 50.28 & 41.62 & 23.63 & 61.90 & 24.93 & 35.65 & 23.37 \\
LT-MLA & 8 & 57 & 3.324 & 17.66 & 15.20 & 28.62 & 58.38 & 51.78 & 42.68 & \textbf{25.00} & 61.50 & 25.45 & 36.25 & 5.06 \\
LPT-MLA & 8 & 60 & 3.257 & 20.69 & 15.89 & 29.30 & \textbf{59.58} & 50.99 & \textbf{45.37} & 24.83 & 67.10 & \textbf{25.77} & 37.72 & 24.38 \\
\rowcolor{talaRow}
TALA & 10 & 98 & \textbf{3.210} & \textbf{24.14} & \textbf{19.66} & \textbf{29.41} & 58.27 & \textbf{53.75} & 44.40 & \textbf{25.00} & \textbf{69.30} & 25.76 & \textbf{38.85} & \textbf{34.26} \\
\midrule
\multicolumn{15}{l}{\textit{$10^{19}$ FLOPs}} \\
vanilla-LT & 10 & 100 & 3.085 & 24.68 & 19.19 & 31.26 & 60.94 & 50.04 & 44.95 & 25.85 & 66.90 & 25.84 & 38.85 & 40.52 \\
LT-MLA & 12 & 144 & 3.014 & 25.83 & 19.25 & 32.17 & 61.92 & 50.28 & \textbf{47.52} & 25.17 & 67.70 & 26.31 & 39.57 & \textbf{55.23} \\
LPT-MLA & 10 & 99 & 3.016 & 25.98 & 19.54 & 31.86 & 62.19 & \textbf{52.09} & 46.84 & 25.09 & 70.00 & 26.21 & 39.98 & 37.68 \\
\rowcolor{talaRow}
TALA & 12 & 153 & \textbf{2.946} & \textbf{28.74} & \textbf{22.74} & \textbf{34.00} & \textbf{63.55} & 49.88 & 46.97 & \textbf{25.94} & \textbf{70.70} & \textbf{26.78} & \textbf{41.03} & 52.51 \\
\midrule
\multicolumn{15}{l}{\textit{$3\times10^{19}$ FLOPs}} \\
vanilla-LT & 14 & 231 & 2.886 & 27.36 & 23.29 & 35.54 & 64.64 & 50.51 & 48.48 & 26.11 & 74.50 & 26.92 & 41.93 & 50.13 \\
LT-MLA & 14 & 213 & 2.814 & 30.74 & 24.67 & 37.87 & 65.72 & \textbf{52.72} & \textbf{51.68} & \textbf{28.24} & 74.00 & 27.55 & 43.69 & 59.92 \\
LPT-MLA$^{\dagger}$ & 14 & 230 & 2.822 & 31.96 & 23.97 & 36.82 & 64.53 & 51.14 & 49.79 & 27.73 & 72.50 & 27.35 & 42.87 & 53.73 \\
\rowcolor{talaRow}
TALA & 14 & 228 & \textbf{2.760} & \textbf{33.28} & \textbf{27.60} & \textbf{39.48} & \textbf{66.16} & 51.38 & 51.26 & 26.54 & \textbf{75.90} & \textbf{27.96} & \textbf{44.40} & \textbf{63.36} \\
\midrule
\multicolumn{15}{l}{\textit{$10^{20}$ FLOPs}} \\
vanilla-LT & 20 & 606 & 2.674 & 35.69 & 29.44 & 42.66 & 66.59 & 51.70 & 54.12 & 28.07 & 76.60 & 28.74 & 45.96 & 48.18 \\
LT-MLA & 20 & 554 & 2.624 & 37.51 & 31.15 & 45.41 & \textbf{68.34} & 53.28 & 55.30 & \textbf{30.38} & 79.30 & 28.95 & 47.73 & 65.25 \\
LPT-MLA$^{\dagger}$ & 20 & 604 & 2.632 & 37.09 & 28.35 & 44.14 & 67.63 & 51.93 & 54.67 & 28.84 & 77.30 & 29.20 & 46.57 & 68.14 \\
\rowcolor{talaRow}
TALA & 20 & 598 & \textbf{2.591} & \textbf{39.30} & \textbf{32.52} & \textbf{47.19} & 68.28 & \textbf{55.17} & \textbf{55.72} & 30.03 & \textbf{80.40} & \textbf{29.80} & \textbf{48.71} & \textbf{74.81} \\
\bottomrule
}
\end{table}

\noindent\begin{minipage}{\linewidth}
\begin{wrapfigure}{r}{.50\linewidth}
\centering
\includegraphics[width=\linewidth]{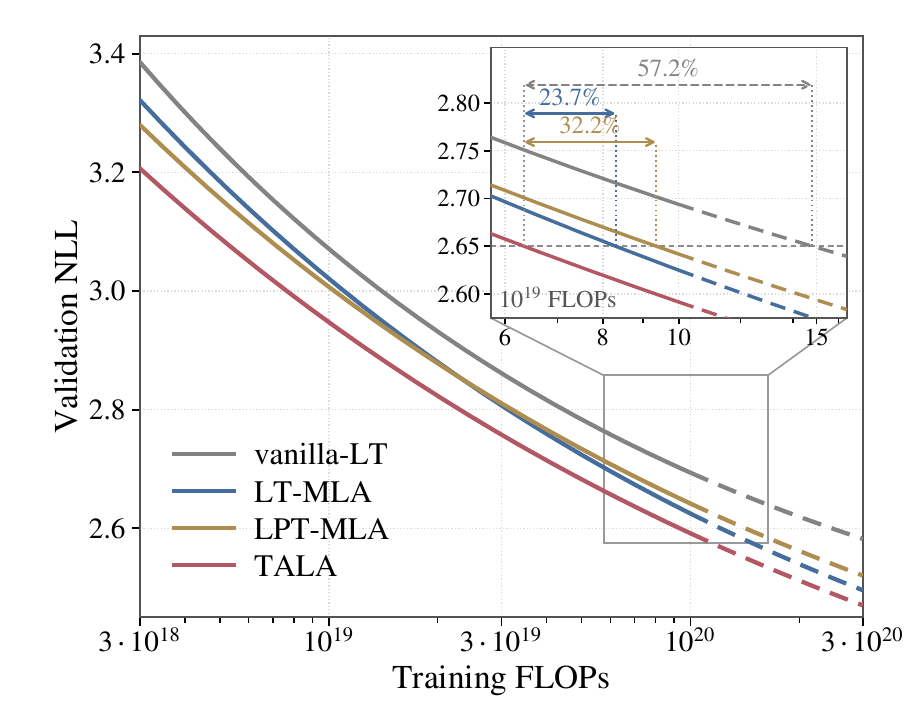}
\caption{\TALAScalingFigureCaption}\label{fig:scaling-frontier}
\end{wrapfigure}
\fi
For each architecture and compute budget, we evaluate the checkpoint with the lowest validation NLL on the downstream benchmarks. As shown in Table~\ref{tab:scaling-selected}, TALA achieves the highest average score on the nine language benchmarks at all four budgets and the highest NIAH score at three of them. Relative to vanilla-LT, LT-MLA, and LPT-MLA, the gains in average language-benchmark scores range from 2.18 to 3.21, 0.71 to 2.60, and 1.06 to 2.14 percentage points, respectively. At the largest compute budget, TALA achieves a mean NIAH score of 74.81\%, exceeding the three reported baseline means by 26.63, 9.56, and 6.67 percentage points, respectively. These results demonstrate that the improvements in validation NLL also translate into stronger downstream performance.

\ifTALAPreprint\else

\fi

\paragraph{Fixed-depth pretraining.}
We further compare the architectures at a fixed physical and recurrent depth, using approximately 450M-parameter models trained on 20B tokens (Table~\ref{tab:native20b}). Compared with vanilla-LT, TALA achieves approximately $12\times$ KV cache compression and uses 28.4\% fewer training FLOPs, while reducing validation NLL by 0.027 and improving the mean language-benchmark and NIAH scores by 1.20 and 6.93 points, respectively. Among the five full-history architectures, TALA achieves the lowest validation NLL and the highest average scores on both benchmark groups. At nearly identical KV capacity, TALA improves both averages over LT-MLA by 1.94 and 5.88 points, respectively. It also outperforms LT-MLA (wide) while using approximately one quarter of its KV cache, improving the mean NIAH score from 63.63\% to 80.74\%.
\ifTALAPreprint
\par\WFclear
\end{minipage}\par
\fi

\newcommand{\TALAFixedDepthGraphic}[1]{%
\centering
\includegraphics[width=\linewidth]{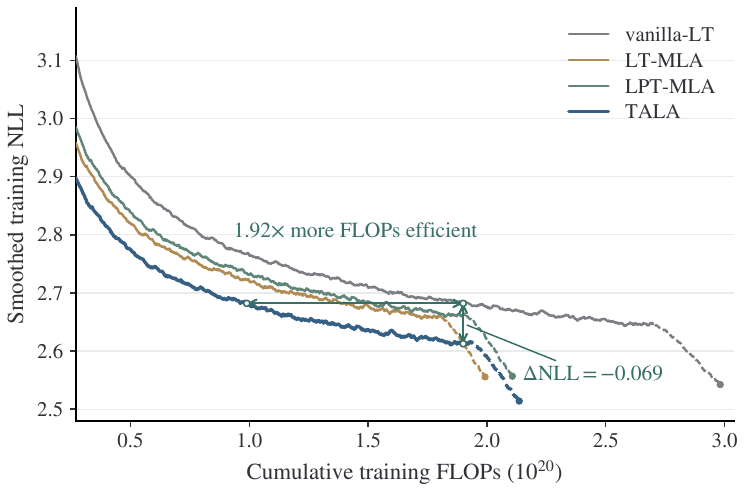}
#1{Training trajectories of looped Transformers ($L=18$, $R=4$, 20B tokens). Final-recursion training NLL is exponentially smoothed ($\beta=0.98$ per logged observation). Solid and dashed segments denote the stable and decay phases. Horizontal and vertical arrows compare TALA with vanilla-LT at matched NLL and matched \mbox{FLOPs}, respectively, within their shared stable phase.}
\label{fig:fixed-depth-training-loss}
}
\ifTALAPreprint\else
\begin{wrapfigure}{R}{.50\linewidth}
\TALAFixedDepthGraphic{\caption}
\end{wrapfigure}
\fi

\ifTALAPreprint
\par
\TALAFloat{table}{!t}{!htbp}
\centering
\caption{Fixed-depth pretraining of approximately 450M-parameter looped Transformers ($L=18$, $R=4$, 20B tokens). KV comp. is persistent BF16 KV compression relative to vanilla-LT at 4K, including positional states. Task abbreviations follow Table~\ref{tab:scaling-selected}. Bold and underlining mark the best and second-best quality values within each group, respectively, where applicable.}\label{tab:native20b}
\TALAtableStyle
\ifTALAPreprint\else\fontsize{8.3}{10}\selectfont\fi
\TALAScoredTable{.15pt}{l!{\hspace{\TALAPreprintValue{1.5pt}{0pt}}}rrr}{11}{4}{%
\toprule
\textbf{Model} & \multicolumn{1}{c}{\textbf{FLOPs}} & \multicolumn{1}{c}{\textbf{KV comp.}} & \multicolumn{1}{c}{\mbox{\textbf{NLL}$\downarrow$}} & \TALAScoreHead{L-O} & \TALAScoreHead{L-S} & \TALAScoreHead{HS} & \TALAScoreHead{PIQA} & \TALAScoreHead{WG} & \TALAScoreHead{A-E} & \TALAScoreHead{A-C} & \TALAScoreHead{SciQ} & \TALAScoreHead{MMLU} & \TALAScoreHead[\uparrow]{Avg.} & \TALAScoreHead[\uparrow]{NIAH} \\
\midrule
\multicolumn{15}{l}{\textit{Without local window}} \\
vanilla-LT & $2.98{\times}10^{20}$ & $1.00\times$ & \underline{2.538} & 40.42 & 32.66 & \underline{49.12} & \textbf{70.02} & 56.04 & 57.66 & \underline{33.11} & \underline{82.10} & 29.82 & \underline{50.11} & 73.81 \\
LT-MLA & $1.99{\times}10^{20}$ & $12.04\times$ & 2.552 & 39.57 & \underline{32.99} & 48.50 & 69.04 & 55.25 & 56.48 & 30.20 & 81.80 & \textbf{30.52} & 49.37 & \underline{74.87} \\
LT-MLA (wide) & $2.16{\times}10^{20}$ & $3.01\times$ & 2.541 & 38.83 & 31.83 & 49.03 & \underline{69.21} & \textbf{57.85} & \underline{59.47} & 31.57 & 81.10 & 30.47 & 49.93 & 63.63 \\
LPT-MLA & $2.11{\times}10^{20}$ & $12.04\times$ & 2.552 & \underline{40.54} & 31.96 & 48.10 & 68.77 & 54.38 & 57.95 & 31.91 & 79.70 & 30.35 & 49.30 & 70.18 \\
\rowcolor{talaRow}
TALA & $2.14{\times}10^{20}$ & $11.93\times$ & \textbf{2.511} & \textbf{41.74} & \textbf{34.87} & \textbf{50.48} & 68.66 & \underline{57.14} & \textbf{60.44} & \textbf{34.30} & \textbf{83.70} & \underline{30.49} & \textbf{51.31} & \textbf{80.74} \\
\midrule
\multicolumn{15}{l}{\textit{With local window (64 tokens)}} \\
LPT-MLA+SWA & $2.24{\times}10^{20}$ & $11.96\times$ & 2.524 & 41.43 & \textbf{35.44} & 50.66 & 69.75 & 55.80 & \textbf{61.62} & \textbf{34.56} & 82.80 & 30.73 & 51.42 & 54.25 \\
\rowcolor{talaRow}
TALA+SWA & $2.19{\times}10^{20}$ & $19.51\times$ & \textbf{2.505} & \textbf{42.32} & 34.76 & \textbf{51.38} & \textbf{70.29} & \textbf{57.06} & 61.53 & 34.39 & \textbf{84.30} & \textbf{31.16} & \textbf{51.91} & \textbf{78.01} \\
\bottomrule
}
\end{table}

\noindent\begin{minipage}[t]{.48\linewidth}
\vspace{0pt}
\fi
\paragraph{Training dynamics.}
Figure~\ref{fig:fixed-depth-training-loss} compares the training trajectories of the four architectures at fixed depth.
TALA achieves lower training NLL than the baselines across the later stages of training at matched training compute. At a matched smoothed training NLL within the shared stable phase, TALA achieves $1.92\times$ the training compute efficiency of vanilla-LT. These training trajectories complement our compute-optimal scaling results and further demonstrate the computational efficiency of token-anchored latent attention.

\paragraph{Compatibility with global--local memory sharing.}
TALA can also be combined with first-recursion history sharing and sliding-window attention. Under the same global--local memory-access policy, TALA+SWA achieves lower validation NLL and improves both downstream averages compared with LPT-MLA+SWA. In particular, it increases the NIAH score from 54.25\% to 78.01\%, while reducing KV storage by 38.7\% and training FLOPs by 2.3\% (Table~\ref{tab:native20b}).

\ifTALAPreprint
\end{minipage}\hfill
\begin{minipage}[t]{.50\linewidth}
\vspace{0pt}
\captionsetup{hypcap=false}
\TALAFixedDepthGraphic{\captionof{figure}}
\end{minipage}\par
\fi

\ifTALAPreprint\else

\fi

\subsection{Scaling Recursion and KV Compression}\label{sec:exp_memory}
\paragraph{Increasing recursion depth.}
We investigate the effect of recursion depth by comparing TALA and LT-MLA with two, four, and eight recursions, keeping both architectures at approximately 12-fold KV compression relative to the corresponding dense model and nearly matching their KV capacities at each depth. Both architectures benefit from additional recursions, while TALA consistently achieves lower validation NLL across all three settings. Its advantage remains substantial and is slightly larger at eight recursions than at two (Figure~\ref{fig:memory-efficiency}a). Notably, TALA with only two recursions achieves a lower validation NLL than LT-MLA with eight recursions (2.985 vs. 3.012), using approximately 71\% less training compute and one quarter of the KV cache at the same training-token budget.

\TALAFloat{figure}{!b}{!htbp}
\centering\includegraphics[width=\linewidth]{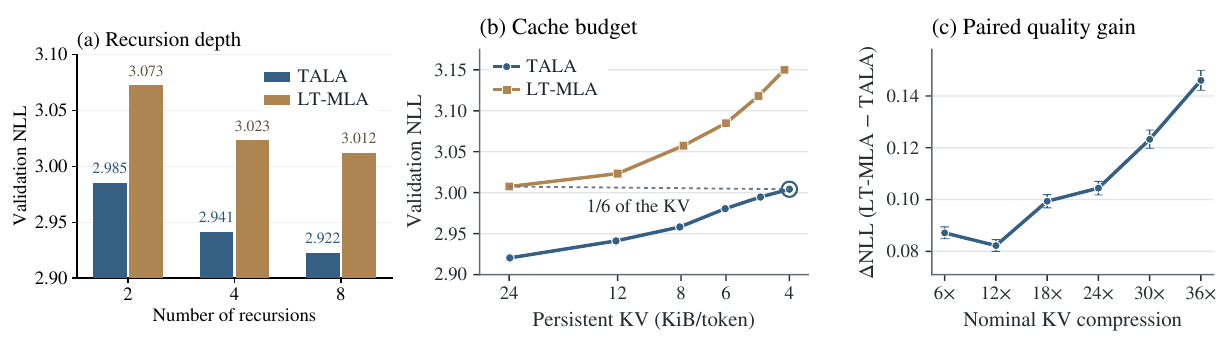}
\caption{Recursion depth and KV compression. (a) Validation NLL by recursion depth (truncated y-axis). (b) Validation NLL versus KV storage. (c) Paired NLL differences with 95\% bootstrap intervals.}\label{fig:memory-efficiency}
\end{figure}

\ifTALAPreprint\TALAFloat{table}{!t}{!htbp}
\centering
\caption{KV compression in non-loop Transformers trained on 10B tokens. Standard models use approximately 450M parameters. GQA-3 and GQA-6 use three and six KV heads, respectively. MLA matches the KV capacity of TALA, while MLA (wide) uses four times this capacity. CLA-2 and CLLA-2 share KV representations and KV latents across pairs of layers, respectively. YOCO-SWA uses sliding-window attention. MemAtten and MHA+KV tables use 1.47B and 2.54B parameters, respectively. KV comp. denotes persistent BF16 KV compression relative to MHA at 4K. Task abbreviations follow Table~\ref{tab:scaling-selected}. Bold and underlining mark the best and second-best quality values within each group, respectively.}\label{tab:nonloop-main}
\TALAtableStyle
\ifTALAPreprint\else\fontsize{8.3}{10}\selectfont\fi
\TALAScoredTable{.16pt}{l!{\hspace{\TALAPreprintValue{1.5pt}{0pt}}}rrr}{11}{4}{%
\toprule
\textbf{Model} & \multicolumn{1}{c}{\textbf{FLOPs}} & \multicolumn{1}{c}{\textbf{KV comp.}} & \multicolumn{1}{c}{\mbox{\textbf{NLL}$\downarrow$}} & \TALAScoreHead{L-O} & \TALAScoreHead{L-S} & \TALAScoreHead{HS} & \TALAScoreHead{PIQA} & \TALAScoreHead{WG} & \TALAScoreHead{A-E} & \TALAScoreHead{A-C} & \TALAScoreHead{SciQ} & \TALAScoreHead{MMLU} & \TALAScoreHead[\uparrow]{Avg.} & \TALAScoreHead[\uparrow]{NIAH} \\
\midrule
\multicolumn{15}{l}{\textit{Standard attention}} \\
MHA & $2.98{\times}10^{19}$ & $1.00\times$ & 2.672 & 33.63 & 26.63 & 41.82 & 67.19 & 52.49 & 53.83 & 28.33 & 74.30 & \underline{29.15} & 45.26 & 58.97 \\
GQA-3 & $2.74{\times}10^{19}$ & $5.62\times$ & 2.699 & 32.91 & 26.04 & 40.69 & 66.27 & 52.80 & 52.57 & 27.99 & 75.60 & 28.46 & 44.81 & 42.94 \\
GQA-6 & $2.79{\times}10^{19}$ & $2.92\times$ & 2.691 & 33.15 & 26.28 & 41.15 & 66.54 & 53.59 & 52.74 & 29.10 & 73.70 & 28.96 & 45.02 & 53.81 \\
MLA & $2.79{\times}10^{19}$ & $7.89\times$ & 2.699 & 34.00 & 27.52 & 40.67 & 67.03 & 52.72 & 53.70 & 28.33 & 74.40 & 28.54 & 45.21 & 56.94 \\
MLA (wide) & $3.12{\times}10^{19}$ & $1.97\times$ & 2.684 & 35.07 & 28.51 & 41.54 & 66.16 & 52.96 & 54.34 & 29.10 & \underline{77.20} & 28.55 & 45.93 & 59.60 \\
CLA-2 & $2.83{\times}10^{19}$ & $1.97\times$ & 2.683 & 33.09 & 25.42 & 41.52 & \textbf{67.79} & 51.14 & 53.41 & \underline{29.86} & 76.20 & 28.71 & 45.24 & 43.18 \\
CLLA-2 & $2.86{\times}10^{19}$ & $7.89\times$ & 2.678 & 34.54 & 27.48 & 41.86 & 66.27 & 51.38 & 54.67 & 29.44 & 75.10 & 28.50 & 45.47 & 58.80 \\
YOCO-SWA & $2.72{\times}10^{19}$ & $5.62\times$ & 2.681 & 33.67 & 26.49 & 42.12 & 67.03 & \underline{54.14} & \underline{54.80} & 29.52 & 75.90 & 28.64 & 45.81 & 21.79 \\
\rowcolor{talaRow}
TALA & $2.99{\times}10^{19}$ & $7.89\times$ & \underline{2.654} & \underline{35.80} & \underline{30.16} & \underline{42.59} & 67.30 & 52.72 & \textbf{55.18} & \textbf{29.95} & \textbf{79.00} & 29.09 & \underline{46.87} & \textbf{71.70} \\
\rowcolor{talaRow}
TALA (realloc.) & $3.01{\times}10^{19}$ & $7.89\times$ & \textbf{2.650} & \textbf{37.03} & \textbf{31.50} & \textbf{43.09} & \underline{67.52} & \textbf{55.41} & 53.37 & 29.10 & 76.80 & \textbf{29.28} & \textbf{47.01} & \underline{70.71} \\
\midrule
\multicolumn{15}{l}{\textit{Nominal $16\times$}} \\
MLA-16 & $2.73{\times}10^{19}$ & $15.64\times$ & 2.709 & 32.10 & 27.09 & 40.45 & 66.27 & 50.83 & 52.36 & \textbf{29.01} & 73.50 & 28.46 & 44.45 & 60.65 \\
\rowcolor{talaRow}
TALA-16 & $2.83{\times}10^{19}$ & $15.64\times$ & \textbf{2.673} & \textbf{36.29} & \textbf{27.67} & \textbf{42.38} & \textbf{67.19} & \textbf{53.20} & \textbf{53.24} & \textbf{29.01} & \textbf{76.30} & \textbf{28.97} & \textbf{46.03} & \textbf{71.38} \\
\midrule
\multicolumn{15}{l}{\textit{Nominal $32\times$}} \\
MLA-32 & $2.71{\times}10^{19}$ & $32.85\times$ & 2.749 & 30.35 & 22.41 & 39.06 & 66.32 & 48.93 & \textbf{52.86} & 27.65 & 69.50 & \textbf{28.08} & 42.80 & 36.77 \\
\rowcolor{talaRow}
TALA-32 & $2.74{\times}10^{19}$ & $32.85\times$ & \textbf{2.709} & \textbf{33.15} & \textbf{27.42} & \textbf{40.50} & \textbf{66.59} & \textbf{52.64} & 51.14 & \textbf{27.82} & \textbf{72.50} & 27.80 & \textbf{44.40} & \textbf{52.00} \\
\midrule
\multicolumn{15}{l}{\textit{Token-table baselines}} \\
MemAtten & $2.83{\times}10^{19}$ & $1.00\times$ & 2.623 & 34.93 & \textbf{29.48} & 43.91 & 68.17 & \textbf{53.04} & \textbf{54.84} & \textbf{30.29} & 74.70 & \textbf{29.68} & \textbf{46.56} & 48.81 \\
MHA+KV tables & $2.98{\times}10^{19}$ & $1.00\times$ & \textbf{2.606} & \textbf{35.77} & 29.42 & \textbf{44.36} & \textbf{68.72} & 50.91 & 54.34 & 29.18 & \textbf{76.60} & 29.42 & 46.52 & \textbf{63.68} \\
\bottomrule
}
\end{table}
\fi

\paragraph{Increasing KV compression.}
With the number of recursions fixed at four, we further evaluate TALA and LT-MLA across six nominal KV compression ratios, ranging from $6\times$ to $36\times$. As shown in Figure~\ref{fig:memory-efficiency}b,c, TALA consistently outperforms LT-MLA at nearly identical cache capacities, with its advantage becoming more pronounced under tighter cache budgets. Remarkably, TALA at $36\times$ compression achieves slightly lower validation NLL than LT-MLA at $6\times$ compression, demonstrating that comparable language modeling quality can be achieved with approximately one sixth of the KV cache required by LT-MLA. These results highlight the effectiveness of shared token latents under increasingly restrictive memory budgets.

\subsection{KV Compression in Non-Loop Transformers}\label{sec:exp_nonloop}
To investigate whether the benefits of token-anchored latent attention extend beyond recurrent computation, we apply TALA to non-loop Transformers ($R=1$), where a single base is shared across all physical layers. We compare approximately 450M-parameter models trained on 10B tokens against eight standard attention baselines, including dense MHA, GQA~\citep{gqa_2023}, independent MLA, CLA and CLLA~\citep{cla_2024,clla_2024}, and YOCO-SWA~\citep{yoco_2024}. The specific baseline configurations are summarized in Table~\ref{tab:nonloop-main}. Since the first physical layer receives a token-level input, we additionally evaluate TALA (realloc.), which removes the stage of the first layer and redistributes its capacity to the base and remaining stages while maintaining the same total KV capacity.

\paragraph{Comparison with existing architectures.}
As shown in Table~\ref{tab:nonloop-main}, TALA achieves lower validation NLL and higher downstream averages than all eight standard attention baselines. At identical KV capacity, TALA improves the average language-benchmark score over MLA and CLLA-2 by 1.66 and 1.40 percentage points, respectively, with corresponding NIAH gains of 14.76 and 12.90 percentage points. Compared with dense MHA, TALA improves both downstream averages by 1.61 and 12.73 percentage points while using approximately one eighth of its KV cache and nearly identical training compute. It also outperforms MLA (wide) with one quarter of its KV cache. Reallocating the first-layer stage further improves validation NLL and the average language-benchmark score, while the default TALA achieves the highest NIAH score among the standard comparisons.

\ifTALAPreprint\else\fi

\paragraph{Higher KV compression.}
The advantage of TALA over independent MLA persists under more restrictive cache budgets. At nominal $16\times$ and $32\times$ compression, TALA improves the average language-benchmark scores by 1.58 and 1.60 percentage points, and the NIAH scores by 10.73 and 15.23 percentage points, respectively. These results extend our observations in Section~\ref{sec:exp_memory} to non-loop architectures, highlighting the effectiveness of shared token representations under tight KV budgets.

\newcommand{\TALAMatchingGraphic}{%
\centering\includegraphics[width=0.96\linewidth]{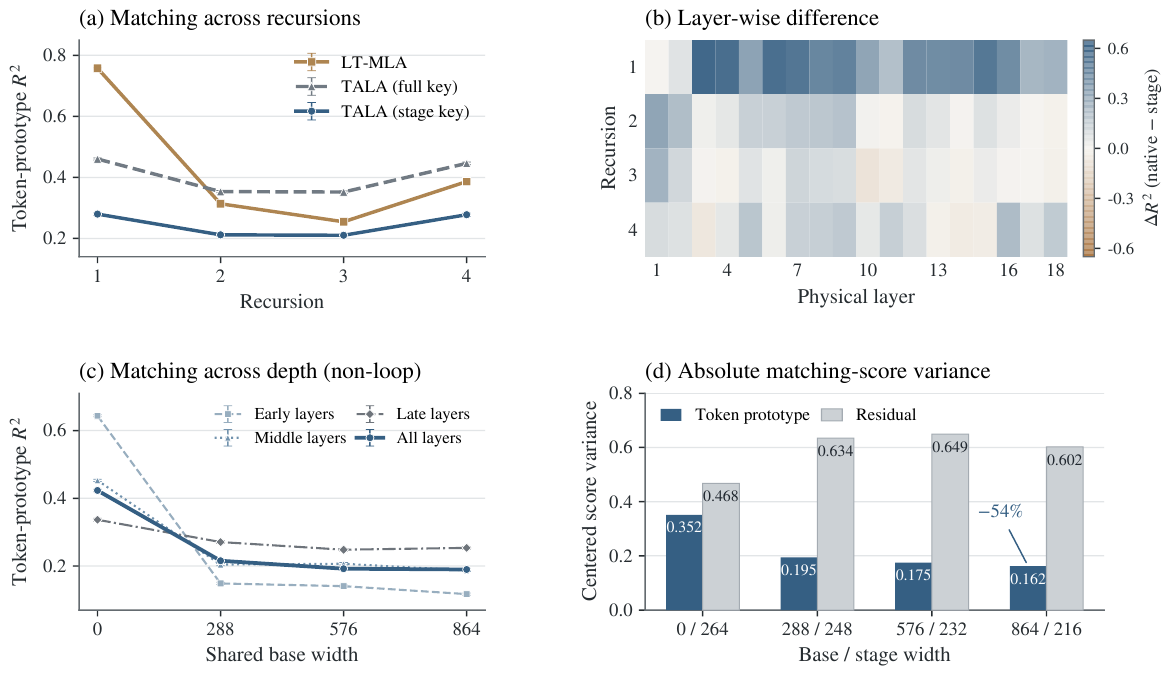}
\caption{Functional specialization in attention matching. (a) Token-prototype $R^2$ for LT-MLA native keys and TALA full and stage keys. (b) Layer-wise difference $R^2_{\mathrm{native}}-R^2_{\mathrm{stage}}$. (c) Non-loop stage-key predictability at fixed KV capacity. (d) Corresponding prototype and residual score variances. Error bars in (a,c) denote 95\% document-bootstrap intervals.}\label{fig:mechanism}
}
\ifTALAPreprint
\begin{figure}[!t]
\TALAMatchingGraphic
\end{figure}
\fi

\paragraph{Comparison with learned token tables.}
We further compare TALA with our implementation of Memory Attention~\citep{kang2026memoryattention}, denoted MemAtten in Table~\ref{tab:nonloop-main}, which constructs values by adding learned token-indexed representations to contextual keys. To examine the benefits of explicitly incorporating token-specific features in both keys and values, we also construct an MHA variant, denoted MHA+KV tables, that adds independently learned token-indexed key and value vectors to the ordinary contextual keys and values at each physical layer. Each layer has its own two lookup tables. This variant improves validation NLL over dense MHA from 2.672 to 2.606, indicating the potential value of providing token-specific features directly to attention. Both table-based models attain lower validation NLL than TALA, but require 1.47B and 2.54B parameters, respectively, and approximately eight times its KV capacity. TALA nevertheless achieves higher downstream averages than both, with a mean NIAH score of 71.70\% compared with 48.81\% and 63.68\%, respectively. These comparisons illustrate the quality--resource trade-off of sharing a compact representation derived from existing token embeddings without introducing additional learned token tables.

\paragraph{Longer pretraining.}
We extend four representative architectures to 20B training tokens. TALA maintains its advantage over MLA at identical KV capacity, improving the average language-benchmark score by 1.34 percentage points and the NIAH score from 60.82\% to 75.81\%. It also outperforms dense MHA and MHA with direct token-dependent KV projections on validation NLL and both downstream averages. Detailed 20B results are provided in Appendix~\ref{app:nonloop_20b}.

\subsection{Functional Specialization in Attention Matching}\label{sec:exp_mechanism}
The shared base in TALA provides token-specific features to all layer executions. We hypothesize that access to these features reduces the need for stage-specific representations to repeatedly encode token-dependent information, allowing their capacity to be used for contextual refinement. To investigate this hypothesis, we examine the matching functions learned by the base and stage keys.

\ifTALAPreprint\else
\begin{figure}[!htbp]
\TALAMatchingGraphic
\end{figure}
\fi

\paragraph{Measuring token-dependent matching.}
For a fixed physical layer, recursion, and attention head, let $x_j$ denote the identity of source token $j$, and $\mathbf{k}_j^S$ its decoded stage key. We partition the analysis documents into disjoint calibration and test sets and construct a token prototype $\boldsymbol{\mu}^S(x)$ by averaging the stage keys of occurrences of token $x$ in the calibration documents. We evaluate on held-out test documents using source tokens with available prototypes. Keeping the realized query and legal source positions fixed, we compare the actual content attention score with the score obtained by replacing the stage key with its corresponding token prototype:
\begin{equation}
    a_{ij}^{S}=\frac{(\mathbf{q}_i^C)^\top\mathbf{k}_j^S}{\sqrt{d_a}},\qquad
    \widehat{a}_{ij}^{S}=\frac{(\mathbf{q}_i^C)^\top\boldsymbol{\mu}^S(x_j)}{\sqrt{d_a}}.
    \label{eq:exp_token_prototype}
\end{equation}
Here, $d_a$ is the attention-score scaling dimension. Both scores exclude positional contributions. We center the actual and prototype-based scores separately over the legal source positions of each query, obtaining $\widetilde{a}_{ij}^{S}$ and $\widetilde{\widehat{a}}_{ij}^{S}$, respectively. We quantify token-dependent matching using the out-of-sample coefficient of determination:
\begin{equation}
    R_{\mathrm{tok}}^2=1-\frac{\sum_{i,j}(\widetilde{a}_{ij}^{S}-\widetilde{\widehat{a}}_{ij}^{S})^2}{\sum_{i,j}(\widetilde{a}_{ij}^{S})^2}.
    \label{eq:exp_token_r2}
\end{equation}
A higher $R_{\mathrm{tok}}^2$ indicates that attention scores are more predictable from source token identity, conditional on the realized queries; negative values are possible. We apply the same procedure to the native content keys of LT-MLA and the full content keys of TALA. The base keys are fully predictable from token identity by construction. All measurements use frozen checkpoints and a frequency-matched shuffled-token control. In the non-loop analysis, the calibration/test partition contains 215/843 documents, and requiring at least 32 calibration occurrences yields 685 eligible token identities. We evaluate 1,955 fixed queries, each with up to 256 legal source positions (Appendix~\ref{app:exp_mechanism_details}).

\paragraph{Functional specialization across recursions.}
We analyze TALA and LT-MLA models trained on 20B tokens with nearly identical KV capacities. As shown in Figure~\ref{fig:mechanism}a, the stage keys of TALA exhibit consistently lower token-prototype predictability across all four recursions. For example, $R_{\mathrm{tok}}^2$ decreases from 0.758 to 0.280 in the first recursion and from 0.386 to 0.278 in the fourth. Meanwhile, the full keys of TALA exhibit higher predictability than LT-MLA in all three subsequent recursions. The layer-wise analysis in Figure~\ref{fig:mechanism}b reveals a similar pattern across most layer executions. These observations suggest that the shared base reduces the need for stage keys to reproduce token-dependent matching while preserving such matching in the combined representation.

\paragraph{Functional specialization across layers.}
To further investigate the role of the shared base independently of recurrence, we train approximately 450M-parameter non-loop Transformers on 10B tokens, varying the base and stage widths while keeping the total KV capacity fixed. As the base width increases from 0 to 864, the overall stage-key $R_{\mathrm{tok}}^2$ decreases from 0.423 to 0.190, with the same trend observed across early, middle, and late layers (Figure~\ref{fig:mechanism}c).

Moreover, the absolute variance of token-prototype scores decreases by approximately 54\%, from 0.352 to 0.162, while the residual score variance increases from 0.468 to 0.602 (Figure~\ref{fig:mechanism}d). This reduction in token-predictable matching occurs alongside improved validation NLL, despite the decreasing stage width.

Together, these observations support a functional specialization in which the shared base provides reusable token-dependent matching features, while the compact stage is less reliant on reproducing these features. The residual may also contain unexplained token interactions and noise. Detailed measurement protocols and additional layer-wise and head-wise analyses are provided in Appendix~\ref{app:exp_mechanism_details}.

\ifTALAPreprint\ifTALAPreprint
\noindent\begin{minipage}{\linewidth}
\captionsetup{type=table}
\else
\begin{table}[!htbp]
\fi
\centering
\caption{Stage allocation and projection sharing at 20B tokens. Nonuniform and uniform variants use stage-width ratios of approximately $0\!:\!2\!:\!1\!:\!1$ and $1\!:\!1\!:\!1\!:\!1$, respectively. Task abbreviations follow Table~\ref{tab:scaling-selected}. Bold marks the best quality value within each model family.}
\label{tab:ablation-main}
\TALAtableStyle
\ifTALAPreprint\else\fontsize{8.3}{10}\selectfont\fi
\TALAScoredTable{.16pt}{l!{\hspace{\TALAPreprintValue{1.5pt}{0pt}}}rr}{11}{3}{%
\toprule
\textbf{Model} & \multicolumn{1}{c}{\textbf{FLOPs}} & \multicolumn{1}{c}{\TALAMin{NLL}} & \TALAScoreHead{L-O} & \TALAScoreHead{L-S} & \TALAScoreHead{HS} & \TALAScoreHead{PIQA} & \TALAScoreHead{WG} & \TALAScoreHead{A-E} & \TALAScoreHead{A-C} & \TALAScoreHead{SciQ} & \TALAScoreHead{MMLU} & \TALAScoreHead[\uparrow]{Avg.} & \TALAScoreHead[\uparrow]{NIAH} \\
\midrule
\multicolumn{14}{l}{\textit{TALA variants}} \\
TALA (nonuniform, untied) & $2.13{\times}10^{20}$ & 2.512 & 41.24 & 36.50 & 50.35 & \textbf{69.31} & 55.17 & 58.96 & 33.87 & 83.60 & 30.32 & 51.04 & 75.17 \\
TALA (untied) & $2.14{\times}10^{20}$ & \textbf{2.504} & \textbf{42.69} & \textbf{37.47} & \textbf{50.80} & \textbf{69.31} & 55.88 & 58.59 & 31.23 & 83.60 & \textbf{30.59} & 51.13 & 73.73 \\
\rowcolor{talaRow}
TALA & $2.14{\times}10^{20}$ & 2.511 & 41.74 & 34.87 & 50.48 & 68.66 & \textbf{57.14} & \textbf{60.44} & \textbf{34.30} & \textbf{83.70} & 30.49 & \textbf{51.31} & \textbf{80.74} \\
\midrule
\multicolumn{14}{l}{\textit{LT-MLA variants}} \\
LT-MLA & $1.99{\times}10^{20}$ & 2.552 & 39.57 & 32.99 & 48.50 & 69.04 & \textbf{55.25} & 56.48 & 30.20 & \textbf{81.80} & \textbf{30.52} & 49.37 & \textbf{74.87} \\
LT-MLA (untied) & $1.99{\times}10^{20}$ & \textbf{2.547} & \textbf{40.00} & \textbf{34.19} & \textbf{48.55} & \textbf{69.37} & 54.62 & \textbf{57.74} & \textbf{30.80} & 80.70 & 30.10 & \textbf{49.56} & 68.12 \\
\bottomrule
}
\ifTALAPreprint
\end{minipage}\par
\else
\end{table}
\fi
\fi
\subsection{Architectural Ablations}\label{sec:exp_ablation}
\paragraph{Stage allocation.}
We investigate how stage capacity should be distributed across recursions by comparing our earlier nonuniform allocation, with stage widths in an approximate ratio of $0:2:1:1$, with the uniform allocation of $1:1:1:1$ used in TALA. The former omits the first-recursion stage and allocates more capacity to the second recursion; the latter assigns an equally sized stage to every recursion. Both variants use independent stage projections across recursions and have comparable KV capacities. As shown in Table~\ref{tab:ablation-main}, the uniform allocation reduces validation NLL from 2.512 to 2.504, with broadly comparable downstream performance. We adopt uniform stage widths in our final architecture, allowing every layer execution to complement the shared base with its own stage representation.

\ifTALAPreprint\else\fi

\paragraph{Sharing projection parameters.}
We further examine whether stage projections need to be specialized across recursions. Sharing these parameters preserves the dimensions and number of stored stage latents while reducing the model's parameter count. We compare shared and independent projections at identical latent widths for both TALA and LT-MLA.

Although independent projections achieve slightly lower validation NLL in both architectures, parameter sharing reduces the parameter count of TALA from approximately 479M to 446M and improves its NIAH score from 73.73\% to 80.74\%. It also achieves the highest average language-benchmark score among the TALA variants. We therefore share projection parameters across recursions in our main architecture, preserving the parameter-sharing principle of looped Transformers.

\FloatBarrier

\section{Related Work}\label{sec:related_work}

\paragraph{Looped Transformers and recurrent computation.}
Reusing parameters across depth enables neural networks to perform additional computation without increasing the number of unique weights, an idea explored in Universal Transformers and ALBERT~\citep{universal_transformer_2019,lan2020albert}. Recent works, such as Huginn and Ouro, revisit recurrent depth in language modeling~\citep{huginn_2025,ouro_2025}, with subsequent studies investigating recurrence topology, recurrent-state updates, and execution schedules~\citep{huang2026loopedright,plt_2025,asymloop_2026}. The recent adoption of recurrent computation in models such as Inkling 2 Mini and GLM-5.4 further highlights the importance of efficient looped architectures~\citep{thinkingmachines2026inkling2mini,zai2026glm54}.

\paragraph{KV-cache efficiency in looped Transformers.}
Existing approaches explore KV efficiency at different stages of model development. Inference-time methods exploit redundancy in pretrained models: FlashLoop combines sparse updates with quantized KV residuals, LoopKV allocates precision and reuse policies according to loop--layer sensitivity, and LoopVQ jointly quantizes KV representations across recursions~\citep{flashloop_2026,loopkv_2026,loopvq_2026}.

Adaptation-based approaches further improve the compressibility or shareability of KV representations. Looped Latent Attention (LLA) compresses low-rank KV trajectories across recursions, while Hybrid Latent Attention (HLA) combines compressed historical representations with exact recent KV~\citep{lla_2026,hla_2026}. MELT adapts recurrent attention to a gated shared cache through interpolation and distillation~\citep{melt_2026}. BeNice! combines stochastic-depth pretraining with a post-training objective to enable final-recursion KV sharing~\citep{benice_2026}.

Architectural approaches incorporate memory sharing directly into recurrent computation. PLT and LPT share first-recursion global memory and complement it with local attention~\citep{plt_2025,lpt_2026}. AsymLoop separates single-pass contextual encoding from recurrent refinement, allowing the decoder to jointly attend to fixed global KV and evolving local KV~\citep{asymloop_2026}. GLM-5.4 adopts a global--local memory organization with cross-recursion reuse of sparse KV and attention indices~\citep{zai2026glm54}. TALA explores component-level sharing within each historical KV representation: shared and stage-specific contributions are combined at the same token position, allowing compact contextual refinements throughout the full history. This representation is also compatible with global--local memory sharing.

\paragraph{KV compression and cross-layer sharing.}
Grouped-query attention (GQA) reduces KV storage by sharing keys and values across groups of query heads~\citep{gqa_2023}, while multi-head latent attention (MLA) compresses KV into low-dimensional latents~\citep{deepseekv2_2024}. Cross-Layer Attention (CLA) shares KV across neighboring layers, and Cross-Layer Latent Attention (CLLA) extends this principle to latent representations with layer-specific projections~\citep{cla_2024,clla_2024}. YOCO separates KV construction from its reuse through a decoder--decoder architecture~\citep{yoco_2024}. TALA builds on latent compression and cross-layer sharing by constructing a global latent before contextual computation and complementing it with a compact, independent latent at each layer execution.

\paragraph{Token-indexed memory and functional specialization.}
Recent studies investigate how explicit access to token-dependent information can improve contextual computation. Per-Layer Embeddings (PLE) in Gemma models supply token-indexed features to individual layers~\citep{google2025gemma3n,gemmateam2026gemma4}. STEM replaces the FFN up-projection with token-indexed lookup, while Engram introduces conditional n-gram memory to alleviate repeated reconstruction of lexical information~\citep{sadhukhan2026stem,cheng2026engram}. Chorus further explores sharing n-gram embedding tables across layers through layer-specific adaptation~\citep{chung2026chorus}. In attention, Bank of Values (BoV) introduces context-free, token-indexed values in deeper layers, while Memory Attention combines contextual keys with learned token memory to construct values~\citep{he2026bov,kang2026memoryattention}. TALA derives a shared latent directly from existing token embeddings, without introducing additional token memory tables, and uses it to construct both keys and values across layers and recursions. Our work further investigates how reusable token-specific matching functions can be absorbed by shared representations, freeing stage-specific latent capacity for contextual information.

\section{Conclusion}\label{sec:conclusion}

In this work, we investigate how KV memory can be efficiently organized across successive computations in looped Transformers. Our findings motivate Token-Anchored Latent Attention (TALA), which shares a single token-derived base latent across all layers and recursions, complemented by compact stage-specific latents. \textbf{\boldmath Compared with vanilla looped Transformers in equal-token pretraining, TALA achieves approximately $12\times$ KV cache compression and uses 28.4\% fewer training FLOPs, while attaining lower validation NLL and higher downstream benchmark scores.} TALA also achieves $1.92\times$ the training compute efficiency of vanilla looped Transformers at a matched training NLL, with consistent advantages in compute-optimal scaling, across recursion depths, and under aggressive KV compression. The benefits further extend to conventional Transformers. Our mechanistic analyses suggest that shared token latents facilitate functional specialization by absorbing reusable token-specific matching functions and freeing stage-specific representational capacity for contextual information. This perspective opens up broader opportunities for architectural design, including adaptive memory allocation across computational depth, more memory-efficient scaling of recurrent computation, and shared latent memory in multimodal architectures. We believe that exploiting the distinction between reusable and computation-specific information offers a promising direction for improving the efficiency and expressivity of future neural architectures.

\bibliography{references}
\bibliographystyle{iclr2027_conference}

\appendix
\TALAAppendixLayout
\raggedbottom
\section{Architecture and Memory Organization}\label{app:memory_details}
\subsection{Causal Writers, Readouts, and Cache Accounting}

\paragraph{Causal writers and readouts.}
Let $G$ be the number of consecutive physical layers per group, where $G$ divides $L$, with $g(\ell)=\lceil\ell/G\rceil$ the group of layer $\ell$ and $\ell_g=(g-1)G+1$ its first layer. For source position $j$, $\mathbf h_{\ell,j}^{(r)}$ is the pre-attention state at recursion $r$. Each group writes its base before attention in layer $\ell_g$ of recursion one, and each layer execution writes its stage.
\begin{equation}\label{eq:app_grouped_write}
\begin{aligned}
\mathbf b_{g,j}&=W_{g,B}^{DKV}\mathbf h_{\ell_g,j}^{(1)},\\
\mathbf s_{\ell,j}^{(r)}&=W_{\ell,S}^{DKV}\mathbf h_{\ell,j}^{(r)},
  \quad r=1,\ldots,R,
\end{aligned}
\end{equation}
Each reader decodes the base $\mathbf b_{g(\ell),j}$ and its current stage $\mathbf s_{\ell,j}^{(r)}$ at the same source position using layer-specific projections (Eqs.~\eqref{eq:group_base} and~\eqref{eq:base_stage_decode}). Full-stack sharing ($G=L$) uses one base derived from the initial token state $\mathbf h_j^{(0)}=\mathbf h_{1,j}^{(1)}$. Projection weights are shared across recursions, while stage states are specific to each layer execution.

\paragraph{Persistent KV storage.}
Decoded content keys accompany decoupled RoPE positional keys. TALA stores one width-$d_p=32$ positional state per token and physical layer, shared across recursions. For scalar size $\eta$ bytes (two for BF16) and source length $T$, the grouped cache size is
\begin{equation}\label{eq:app_persistent_memory}
M_G(T)=\eta T\left(\frac{L}{G}d_B+LRd_S+Ld_p\right).
\end{equation}
Under full-stack sharing, the cache per token and physical layer is $\eta(d_B/L+Rd_S+d_p)$. For $L=18$, $R=4$, and $(d_B,d_S)=(704,176)$, this is approximately 1,550.2 bytes. KV storage counts persistent attention states and excludes temporary or reconstructed tensors.

\paragraph{Global--local retention.}
At recursion $r>1$, a query at position $i$ uses the first-recursion global history and the current recursion's causal window of width $w$. Suppressing the physical-layer index, its candidate slots are
\[
\{(1,j):j\leq i\}\ \cup\ \{(r,j):j\leq i,\ i-j<w\}.
\]
Both sets are restricted to the same document. Each slot uses its source token's base and its donor recursion's stage through Eq.~\eqref{eq:base_stage_decode}. The two branches share one softmax and overlapping positions remain distinct candidates. The first recursion stores the full compressed stage history with its positional states. The local branch stores decoded content keys and values and their positional states. For width $d$ of each content key or value tensor, the persistent storage per token and physical layer is
\[
\frac{M_{\mathrm{hybrid}}(T)}{LT}
=\eta\left(\frac{d_B}{L}+d_S+d_p
 +(R-1)\frac{\min(w,T)}{T}(2d+d_p)\right).
\]
The first three terms account for the shared base and global stage and positional states. The last accounts for the decoded local keys and values and positional states across later recursions. The identities in Section~\ref{sec:method_mechanism} apply to these fixed candidate slots.

TALA+SWA uses $(d_B,d_S,w)=(1088,272,64)$, $L=18$, $R=4$, $d=1152$, and $d_p=32$. At $T=4096$, the BF16 local term contributes 219 bytes per token per physical layer, giving 947.8889 bytes in total, or 17,062 bytes per token for the whole model. This corresponds to $19.51\times$ compression relative to vanilla-LT.

\paragraph{Parameters and compute.}
We count unique parameters once and include all executed recursions in the forward and backward FLOPs. Each base is constructed once per source token and sharing group.

\paragraph{Stage-key score rank.}
For a fixed physical layer, recursion, and head, let $Q^C\in\mathbb R^{N_q\times d_h^C}$ denote content queries, $S\in\mathbb R^{N_k\times d_S}$ source stage latents, and $U_S^K\in\mathbb R^{d_h^C\times d_S}$ the head-specific stage-key decoder. The unmasked linear stage contribution to the content-score matrix is
\begin{equation}\label{eq:app_rank_score}
\Delta A=\frac{Q^C U_S^K S^\top}{\sqrt{d_a}},\qquad
\operatorname{rank}(\Delta A)\le d_S.
\end{equation}
Here $d_h^C$ is the per-head content dimension $d_h$ from Eq.~\eqref{eq:prelim_mla}, and $d_a$ is the native score-scaling dimension.

\paragraph{Attention identities.}
With queries, positional scores, and legal source positions fixed, the full softmax normalizer is $Z_F=Z_B\sum_j p_B(j)e^{\delta_j}$, which yields Eq.~\eqref{eq:stage_exponential_tilt}. Expanding $O_F=A_F(V_B+\Delta V)$ and subtracting $O_B=A_BV_B$ gives Eq.~\eqref{eq:stage_value_identity} before the common output projection.

\paragraph{Shared linear readouts.}
Suppose a common latent $c=Bx\in\mathbb R^m$ supports $G$ linear readouts $A_\ell x=U_\ell c$. With unrestricted linear maps, exact reconstruction for all $x$ is possible if and only if
\begin{equation}\label{eq:shared_linear_rank_condition}
m\ge\operatorname{rank}\!\begin{bmatrix}
A_1\\[-1mm]\vdots\\[-1mm]A_G
\end{bmatrix}.
\end{equation}
Stacking the readouts gives $A=UB$, so $\operatorname{rank}(A)\leq m$ is necessary. Conversely, a rank factorization of $A$ supplies $U$ and $B$, with zero padding when $m>\operatorname{rank}(A)$.

\subsection{Memory-Reuse Diagnostic}\label{app:soft_budget}
The memory-reuse diagnostic uses full-history MHA with four recursions and the final-recursion next-token prediction objective. Token-independent key and value mixtures are learned separately for each layer, head, and reader over the legal writers in Section~\ref{sec:method_discovery}. Banks are mixed at each source position before one causal attention operation. We compute scale-adjusted usage coefficients from the RMS magnitudes of valid source tokens. For two approximately 150M-parameter models trained on 10B tokens, increasing $\lambda$ from 0 to 0.04 changes validation NLL from 2.785 to 2.784 and reduces the soft bank-count proxy from 3.412 to 2.512. All four banks are retained under simulated storage pressure.

\subsection{Sharing Span}\label{sec:sharing-span-details}
Table~\ref{tab:sharing_span} reports the sweep underlying Figure~\ref{fig:tala-sharing-span}. All five models use 12 physical layers, four recursions, 3B training tokens, base/stage widths of 448/112, and a width-32 positional state. The span $G$ varies from one layer to the entire stack, with a stage at every layer execution and projection weights shared across recursions.

\TALAFloat{table}{!htbp}{!ht}
\centering
\ifTALAPreprint\captionsetup{width=.80\linewidth}\fi
\caption{Base-sharing span sweep ($L=12$, $R=4$, 3B tokens, $d_B=448$, $d_S=112$). KV is whole-model persistent BF16 storage per token.}
\label{tab:sharing_span}
\TALATableStyle
\begin{tabular}{@{}rrrrrr@{}}
\toprule
$G$ & Bases & Params (M) & Train (EF) & KV (B/token) & \TALAMin{NLL} \\
\midrule
1 & 12 & 157 & 10 & 22272 & 2.982 \\
2 & 6 & 155 & 10 & 16896 & 2.954 \\
4 & 3 & 154 & 10 & 14208 & 2.950 \\
6 & 2 & 154 & 10 & 13312 & \underline{2.943} \\
12 & 1 & 153 & 10 & 12416 & \textbf{2.941} \\
\bottomrule
\end{tabular}
\end{table}

\section{Training and Looped-Transformer Experiments}\label{app:exp_protocol}
\subsection{Training and Evaluation Details}\label{app:exp_training_details}

\paragraph{Training.}
The recursion-depth and compression sweeps, fixed-depth looped models, and non-loop Transformers use GPT-2 tokenization, sequence length 4,096, and packed-document causal attention with position resets. We use AdamW with $(\beta_1,\beta_2)=(0.9,0.95)$, $\epsilon=10^{-8}$, weight decay 0.1, and gradient clipping at 1. The global batch contains 524,288 tokens, counted as processed sequence positions. The warmup--stable--decay schedule decays the learning rate from $6\times10^{-4}$ to $6\times10^{-5}$ over the final 10\% of updates. The 3B-token sweeps use 240 warmup steps. The four non-loop 20B extensions inherit a schedule with 264 warmup steps, which is also used for the six additional 10B attention baselines.

\paragraph{Evaluation.}
Validation NLL is averaged over valid next-token targets in 1,058 held-out documents. Downstream evaluation follows Section~\ref{sec:exp_setup}, with six RULER-NIAH tasks at a context length of 4,096. For paired recursion-depth and compression comparisons, we estimate 95\% confidence intervals with 10,000 document-bootstrap replicates, weighting valid next-token targets. Each principal training endpoint uses one initialization seed, and the intervals are conditional on the trained checkpoints.

\paragraph{Fixed-depth looped Transformers.}
We compare models with $L=18$ and $R=4$ trained on 20B tokens. The global--local models use a 64-token local window. Tables~\ref{tab:fixed-resources}--\ref{tab:native20b-niah-full} give configurations, resources, and downstream scores, including the ablations in Section~\ref{sec:exp_ablation}. Figure~\ref{fig:swa-training-loss} shows the training trajectories of the global--local models.
\TALAFloat{table}{!htbp}{!htbp}
\centering
\caption{Configurations and resources for fixed-depth looped models trained on 20B tokens ($L=18$, $R=4$). KV is persistent BF16 bytes per token per physical layer at 4K. An asterisk denotes the first-recursion global latent. The +SWA models retain a 64-token local KV window in later recursions. Proj. denotes cross-recursion projection sharing.}\label{tab:fixed-resources}
\TALAtableStyle
\setlength{\tabcolsep}{.90pt}
\begin{tabular}{lrrrrrrrll}
\toprule
Model & Params (M) & FLOPs (EF) & KV & NLL$\downarrow$ & Avg.$\uparrow$ & NIAH$\uparrow$ & Base & Stage / MLA & Proj. \\
\midrule
\multicolumn{10}{l}{\textit{Without local window}} \\
vanilla-LT & 452.753 & 298.141 & 18496.000 & 2.538 & 50.11 & 73.81 & -- & -- & -- \\
LT-MLA & 414.934 & 199.211 & 1536.000 & 2.552 & 49.37 & 74.87 & -- & 160/160/160/160 & tied \\
LT-MLA (wide) & 450.776 & 216.410 & 6144.000 & 2.541 & 49.93 & 63.63 & -- & 736/736/736/736 & tied \\
LPT-MLA & 450.776 & 210.677 & 1536.000 & 2.552 & 49.30 & 70.18 & -- & 736$^*$ & -- \\
\rowcolor{talaRow}
TALA & 445.937 & 213.561 & 1550.222 & 2.511 & 51.31 & 80.74 & 704 & 176/176/176/176 & tied \\
\midrule
\multicolumn{10}{l}{\textit{With local window}} \\
LPT-MLA+SWA & 444.304 & 223.889 & 1547.000 & 2.524 & 51.42 & 54.25 & -- & 632$^*$ & -- \\
\rowcolor{talaRow}
TALA+SWA & 468.279 & 218.705 & 947.889 & 2.505 & 51.91 & 78.01 & 1088 & 272$^*$ & -- \\
\midrule
\multicolumn{10}{l}{\textit{Additional projection and stage-allocation ablations}} \\
\TALAPreprintValue{TALA (nonuniform, untied)}{\shortstack[l]{TALA (nonuniform,\\untied)}} & 477.954 & 213.341 & 1533.33 & 2.512 & 51.04 & 75.17 & 696 & 0/344/176/176 & untied \\
TALA (untied) & 478.793 & 213.561 & 1550.22 & 2.504 & 51.13 & 73.73 & 704 & 176/176/176/176 & untied \\
LT-MLA (untied) & 444.794 & 199.211 & 1536 & 2.547 & 49.56 & 68.12 & -- & 160/160/160/160 & untied \\
\bottomrule
\end{tabular}
\end{table}

\TALAFloat{table}{!htbp}{!htbp}
\centering
\TALATableStyle
\renewcommand{\arraystretch}{1.10}
\caption{Nine language-benchmark scores (\%) for the fixed-depth 20B looped models and architectural ablations. Task abbreviations follow Table~\ref{tab:scaling-selected}.}\label{tab:native20b-cs-full}
\begin{tabular}{lrrrrrrrrrr}
\toprule
\textbf{Model} & \TALAMax{L-O} & \TALAMax{L-S} & \TALAMax{HS} & \TALAMax{PIQA} & \TALAMax{WG} & \TALAMax{A-E} & \TALAMax{A-C} & \TALAMax{SciQ} & \TALAMax{MMLU} & \TALAMax{Avg.} \\
\midrule
vanilla-LT & 40.42 & 32.66 & \underline{49.12} & \textbf{70.02} & 56.04 & 57.66 & \underline{33.11} & \underline{82.10} & 29.82 & \underline{50.11} \\
LT-MLA & 39.57 & \underline{32.99} & 48.50 & 69.04 & 55.25 & 56.48 & 30.20 & 81.80 & \textbf{30.52} & 49.37 \\
LT-MLA (wide) & 38.83 & 31.83 & 49.03 & \underline{69.21} & \textbf{57.85} & \underline{59.47} & 31.57 & 81.10 & 30.47 & 49.93 \\
LPT-MLA & \underline{40.54} & 31.96 & 48.10 & 68.77 & 54.38 & 57.95 & 31.91 & 79.70 & 30.35 & 49.30 \\
\rowcolor{talaRow}
TALA & \textbf{41.74} & \textbf{34.87} & \textbf{50.48} & 68.66 & \underline{57.14} & \textbf{60.44} & \textbf{34.30} & \textbf{83.70} & \underline{30.49} & \textbf{51.31} \\
\midrule
\multicolumn{11}{l}{\textit{With local window}} \\
LPT-MLA+SWA & 41.43 & 35.44 & 50.66 & 69.75 & 55.80 & 61.62 & 34.56 & 82.80 & 30.73 & 51.42 \\
\rowcolor{talaRow}
TALA+SWA & 42.32 & 34.76 & 51.38 & 70.29 & 57.06 & 61.53 & 34.39 & 84.30 & 31.16 & 51.91 \\
\midrule
\multicolumn{11}{l}{\textit{Additional stage-allocation and projection-sharing ablations}} \\
TALA (nonuniform, untied) & 41.24 & 36.50 & 50.35 & 69.31 & 55.17 & 58.96 & 33.87 & 83.60 & 30.32 & 51.04 \\
TALA (untied) & 42.69 & 37.47 & 50.80 & 69.31 & 55.88 & 58.59 & 31.23 & 83.60 & 30.59 & 51.13 \\
LT-MLA (untied) & 40.00 & 34.19 & 48.55 & 69.37 & 54.62 & 57.74 & 30.80 & 80.70 & 30.10 & 49.56 \\
\bottomrule
\end{tabular}
\end{table}

\TALAFloat{table}{!htbp}{!htbp}
\centering
\caption{Six RULER-NIAH task scores (\%) at 4K for the fixed-depth 20B looped models and architectural ablations. Avg. is the unweighted task mean.}\label{tab:native20b-niah-full}
\TALAtableStyle
\setlength{\tabcolsep}{2pt}
\begin{tabular}{lrrrrrrr}
\toprule
\textbf{Model} & \TALAMax{S1} & \TALAMax{S2} & \TALAMax{S3} & \TALAMax{MK1} & \TALAMax{MV} & \TALAMax{MQ} & \TALAMax{Avg.} \\
\midrule
\multicolumn{8}{l}{\textit{Without local window}} \\
vanilla-LT & 95.00 & 95.00 & \underline{91.40} & 47.60 & \underline{59.70} & 54.15 & 73.81 \\
LT-MLA & \textbf{100.00} & \textbf{99.00} & 87.40 & \underline{50.20} & 55.45 & \underline{57.15} & \underline{74.87} \\
LT-MLA (wide) & \underline{96.60} & 96.40 & 87.80 & 44.60 & 33.55 & 22.80 & 63.63 \\
LPT-MLA & \textbf{100.00} & 94.60 & \textbf{95.60} & 49.20 & 40.45 & 41.20 & 70.18 \\
\rowcolor{talaRow}
TALA & 96.40 & \underline{96.80} & 85.60 & \textbf{76.20} & \textbf{63.00} & \textbf{66.45} & \textbf{80.74} \\
\midrule
\multicolumn{8}{l}{\textit{With local window}} \\
LPT-MLA+SWA & 98.00 & 63.60 & 44.60 & 56.40 & 34.40 & 28.50 & 54.25 \\
\rowcolor{talaRow}
TALA+SWA & 100.00 & 97.60 & 90.00 & 63.60 & 59.90 & 56.95 & 78.01 \\
\midrule
\multicolumn{8}{l}{\textit{Additional stage-allocation and projection-sharing ablations}} \\
TALA (nonuniform, untied) & 99.40 & 97.80 & 87.20 & 77.60 & 45.20 & 43.85 & 75.17 \\
TALA (untied) & 98.40 & 99.80 & 79.40 & 48.20 & 60.20 & 56.40 & 73.73 \\
LT-MLA (untied) & 99.80 & 93.60 & 85.60 & 44.40 & 40.20 & 45.10 & 68.12 \\
\bottomrule
\end{tabular}
\end{table}

\ifTALAPreprint
  \par\noindent\begin{minipage}{\linewidth}
  \captionsetup{hypcap=false,type=figure}
\else
  \begin{figure}[!htbp]
\fi
\centering
\includegraphics[width=\TALAPreprintValue{.54\linewidth}{.72\linewidth}]{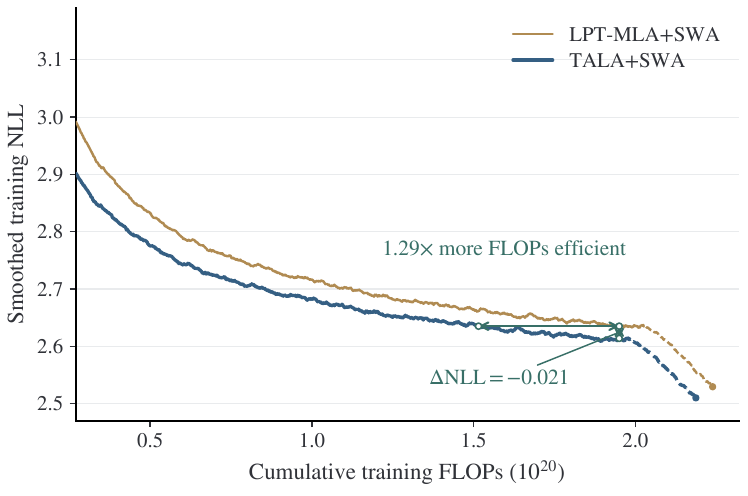}
\caption{Training trajectories with global--local memory sharing (20B tokens, 64-token local window). Smoothing, line styles, and stable-phase comparisons follow Figure~\ref{fig:fixed-depth-training-loss}.}
\label{fig:swa-training-loss}
\ifTALAPreprint
  \end{minipage}\par
\else
  \end{figure}
\fi

\subsection{IsoFLOPs Model Configurations}\label{app:exp_scaling_details}

\paragraph{Model configurations.}
All four architectures use four recursions. The four compute budgets are $3\times10^{18}$, $10^{19}$, $3\times10^{19}$, and $10^{20}$ FLOPs. The corresponding depth grids are $\{6,8,10,12,14,16\}$, $\{8,10,12,14,16,18\}$, $\{12,14,16,18,20,22\}$, and $\{16,18,20,22,24\}$, respectively. Each token budget is set from the model's training FLOPs per token. The three latent architectures have approximately matched KV budgets, and vanilla-LT uses independent dense KV at every recursion. Each architecture contributes 23 model--data allocations, for 92 in total. Tables~\ref{tab:scaling-configurations} and~\ref{tab:scaling-niah-full} report the configurations and individual retrieval scores.
\TALAFloat{table}{!htbp}{!htbp}
\centering
\caption{Model configurations for the IsoFLOPs study ($R=4$). Hidden, FFN, and head counts refer to LT-MLA and TALA. LT-MLA lists the latent width per recursion, LPT-MLA the first-recursion width derived from cache accounting, and TALA the base/stage widths. Parameters are in millions.}\label{tab:scaling-configurations}
\TALAtableStyle
\begin{tabular}{lrrrrrrrrrr}
\toprule
\textbf{Layers} & \textbf{Hidden} & \textbf{FFN} & \textbf{Heads} & \multicolumn{3}{c}{\textbf{Content latent widths}} & \multicolumn{4}{c}{\textbf{Parameters (M)}} \\
\cmidrule(lr){5-7}\cmidrule(l){8-11}
 &  &  &  & \textbf{LT-MLA} & \textbf{LPT-MLA} & \textbf{TALA (B/S)} & \textbf{vanilla-LT} & \textbf{LT-MLA} & \textbf{LPT-MLA} & \textbf{TALA} \\
\midrule
6 & 384 & 1536 & 6 & 32 & 224 & 192/48 & 34 & 32 & 34 & 33 \\
8 & 512 & 2048 & 8 & 56 & 320 & 288/72 & 60 & 57 & 60 & 60 \\
10 & 640 & 2560 & 10 & 72 & 384 & 352/88 & 100 & 93 & 99 & 98 \\
12 & 768 & 3072 & 12 & 96 & 480 & 448/112 & 156 & 144 & 155 & 153 \\
14 & 896 & 3584 & 14 & 120 & 576 & 544/136 & 231 & 213 & 230 & 228 \\
16 & 1024 & 4096 & 16 & 136 & 640 & 608/152 & 329 & 302 & 327 & 323 \\
18 & 1152 & 4608 & 18 & 160 & 736 & 704/176 & 453 & 415 & 451 & 446 \\
20 & 1280 & 5120 & 20 & 184 & 832 & 800/200 & 606 & 554 & 604 & 598 \\
22 & 1408 & 5632 & 22 & 200 & 896 & 864/216 & 791 & 723 & 788 & 779 \\
24 & 1536 & 6144 & 24 & 224 & 992 & 960/240 & 1013 & 924 & 1009 & 998 \\
\bottomrule
\end{tabular}
\end{table}

\paragraph{Loss-surface fitting.}
Let $f=F/10^9$, $d=D/10^{10}$, and $c=C/10^{19}=fd$, with $F$ the training FLOPs per token and $D$ the number of training tokens. We fit a separate loss surface for each architecture:
\begin{equation}
\widehat{\mathcal L}(f,d)=E+A f^{-\alpha}+K d^{-\beta}.
\end{equation}
Let $\theta=(E,\log A,\log K,\alpha,\beta)$ and let $y_i$ be the observed validation NLL. The objective uses Huber loss on log-NLL residuals:
\begin{equation}
\min_\theta\sum_i\rho_{10^{-3}}\!\left(\log\widehat{\mathcal L}_\theta(f_i,d_i)-\log y_i\right).
\end{equation}
We use bounded nonlinear least squares with 48 initializations and select the fit with the lowest objective. Table~\ref{tab:scaling-fit-parameters} reports the coefficients.

\paragraph{Compute-optimal frontier.}
At fixed $c$, substituting $d=c/f$ gives
\begin{equation}
\widehat{\mathcal L}_c(f)=E+A f^{-\alpha}+Kc^{-\beta}f^{\beta}.
\end{equation}
For $A,K,\alpha,\beta>0$, this expression is strictly convex in $\log f$. The first-order condition is $\alpha A f^{-\alpha}=\beta Kc^{-\beta}f^\beta$. Set $q=(\alpha A/(\beta K))^{1/(\alpha+\beta)}$. Then
\begin{equation}
 f^*(c)=q c^{\beta/(\alpha+\beta)},\qquad d^*(c)=q^{-1}c^{\alpha/(\alpha+\beta)}.
\end{equation}
The resulting frontier is
\begin{equation}
\widehat{\mathcal L}^*(C)=E+H(C/10^{19})^{-\gamma},\quad
H=Aq^{-\alpha}+Kq^\beta,\quad \gamma=\frac{\alpha\beta}{\alpha+\beta}.
\end{equation}
At target NLL $y>E$, the fitted compute requirement is $C_y=10^{19}[H/(y-E)]^{1/\gamma}$. TALA's relative compute saving against a baseline is $1-C_{y,\mathrm{TALA}}/C_{y,\mathrm{baseline}}$. Figure~\ref{fig:scaling-frontier} uses solid curves for fitted optima within the convex hull of observed $(\log F,\log D)$ allocations and dashed curves for extrapolation.
\TALAFloat{table}{!htbp}{!htbp}
\centering
\TALATableStyle
\setlength{\tabcolsep}{3.5pt}
\renewcommand{\arraystretch}{1.10}
\caption{Huber loss-surface fits for the four scaling architectures (23 allocations each). $F$, $D$, and $C$ are normalized by $10^9$, $10^{10}$, and $10^{19}$, respectively. The vanilla-LT fit reaches the upper bound on $\beta$ (1.2).}\label{tab:scaling-fit-parameters}
\begin{tabular}{lrrrrrrrr}
\toprule
Architecture & $n$ & $E$ & $A$ & $K$ & $\alpha$ & $\beta$ & $H$ & $\gamma$ \\
\midrule
vanilla-LT & 23 & 2.264080 & 1.008245 & 0.035999 & 0.354361 & 1.200000 & 0.806786 & 0.273574 \\
LT-MLA & 23 & 2.017469 & 0.888153 & 0.236804 & 0.377758 & 0.514708 & 1.003189 & 0.217863 \\
LPT-MLA & 23 & 2.048879 & 0.943913 & 0.114198 & 0.284817 & 0.775885 & 0.957837 & 0.208339 \\
TALA & 23 & 1.955788 & 0.835977 & 0.264435 & 0.350556 & 0.429093 & 0.991418 & 0.192934 \\
\bottomrule
\end{tabular}
\end{table}

\TALAFloat{table}{!htbp}{!htbp}
\centering
\caption{Six RULER-NIAH task scores (\%) at 4K for the selected checkpoints in Table~\ref{tab:scaling-selected}. Avg. is the unweighted task mean. $\dagger$ See Table~\ref{tab:scaling-selected} for the evaluation-sample qualification.}\label{tab:scaling-niah-full}
\TALAtableStyle
\setlength{\tabcolsep}{2pt}
\begin{tabular}{lrrrrrrr}
\toprule
\textbf{Model} & \TALAMax{S1} & \TALAMax{S2} & \TALAMax{S3} & \TALAMax{MK1} & \TALAMax{MV} & \TALAMax{MQ} & \TALAMax{Avg.} \\
\midrule
\multicolumn{8}{l}{\textit{$3\times10^{18}$ FLOPs}} \\
vanilla-LT & 60.20 & 59.80 & 0.60 & 10.80 & 8.80 & 0.00 & 23.37 \\
LT-MLA & 16.00 & 3.20 & 1.00 & 7.20 & 2.90 & 0.05 & 5.06 \\
LPT-MLA & 73.40 & 26.00 & \textbf{4.40} & 19.60 & 13.75 & \textbf{9.15} & 24.38 \\
\rowcolor{talaRow}
TALA & \textbf{82.20} & \textbf{71.60} & 2.80 & \textbf{25.80} & \textbf{19.85} & 3.30 & \textbf{34.26} \\
\midrule
\multicolumn{8}{l}{\textit{$10^{19}$ FLOPs}} \\
vanilla-LT & 90.80 & 85.80 & 34.40 & 21.60 & 9.60 & 0.90 & 40.52 \\
LT-MLA & 95.20 & \textbf{98.60} & \textbf{78.00} & 28.00 & 16.50 & \textbf{15.10} & \textbf{55.23} \\
LPT-MLA & \textbf{98.20} & 56.20 & 11.60 & 27.20 & 21.55 & 11.35 & 37.68 \\
\rowcolor{talaRow}
TALA & 90.60 & 98.40 & 55.40 & \textbf{34.00} & \textbf{21.75} & 14.90 & 52.51 \\
\midrule
\multicolumn{8}{l}{\textit{$3\times10^{19}$ FLOPs}} \\
vanilla-LT & 98.40 & 96.20 & 59.80 & 32.00 & 9.65 & 4.70 & 50.13 \\
LT-MLA & 99.00 & \textbf{100.00} & 87.60 & 29.60 & 21.20 & 22.10 & 59.92 \\
LPT-MLA$^{\dagger}$ & 99.60 & 98.20 & 55.20 & 26.00 & 20.65 & 22.70 & 53.73 \\
\rowcolor{talaRow}
TALA & \textbf{99.80} & 99.20 & \textbf{89.60} & \textbf{39.80} & \textbf{24.20} & \textbf{27.55} & \textbf{63.36} \\
\midrule
\multicolumn{8}{l}{\textit{$10^{20}$ FLOPs}} \\
vanilla-LT & 99.20 & 59.60 & 61.60 & 36.00 & 18.70 & 13.95 & 48.18 \\
LT-MLA & 99.80 & 99.40 & 65.00 & 35.60 & 45.80 & 45.90 & 65.25 \\
LPT-MLA$^{\dagger}$ & \textbf{100.00} & \textbf{99.80} & \textbf{96.20} & 42.60 & 35.75 & 34.50 & 68.14 \\
\rowcolor{talaRow}
TALA & 90.00 & 98.80 & 91.40 & \textbf{54.80} & \textbf{56.70} & \textbf{57.15} & \textbf{74.81} \\
\bottomrule
\end{tabular}
\end{table}

\FloatBarrier

\section{Additional Non-Loop Transformer Experiments}\label{app:exp_nonloop_details}
\subsection{Architecture Variants and Complete 10B Results}\label{app:nonloop_designs}\label{app:nonloop_10b}
All 32 configurations use an 18-layer Transformer with hidden width 1,152, FFN width 4,608, and 18 query heads of width 64, trained on 10B tokens. Table~\ref{tab:nonloop-all} groups the models by architectural design. Tables~\ref{tab:nonloop-commonsense} and~\ref{tab:nonloop-ruler} report individual language and retrieval scores.

\paragraph{Standard attention and cross-layer sharing.}
At matched KV capacity, TALA achieves lower validation NLL and higher downstream averages than MLA and CLLA-2. TALA (realloc.) redistributes the first-layer stage capacity to the base and subsequent stages, improving validation NLL and language-benchmark performance, with slightly lower NIAH than the default TALA.

\paragraph{Compression tiers.}
TALA improves over cache-matched MLA at both nominal 16-fold and 32-fold compression. At the 16-fold tier, it also outperforms MLA with four times the cache on all three quality metrics. At the 32-fold tier, the wider MLA performs better on these metrics.

\paragraph{Base/stage capacity allocation.}
We vary the base/stage widths over $0/264$, $288/248$, $576/232$, and $864/216$ at a fixed KV capacity of 10,656 bytes per token. Validation NLL improves monotonically as the base width increases, and NIAH peaks at the $576/232$ allocation. Appendix~\ref{app:exp_mechanism_details} analyzes the matching functions of these models.

\paragraph{Token tables and direct token readouts.}
The following variants test whether token identity should enter the keys, the values, or both, and whether to represent it with a learned table or a projection of the existing input embedding. Let $x_j$ be the source token ID and $\mathbf e_j$ its input embedding.

\textbf{MemAtten} constructs the value from the contextual content key before positional encoding, plus a layer-specific token lookup with per-head RMS normalization. It replaces the independent contextual value projection. \textbf{MHA+KV tables} retains both standard MHA projections and adds a separate normalized lookup to each path. The key addition precedes positional encoding, and neither table is shared across layers. \textbf{MHA+VTable} and \textbf{MHA+KTable} keep the same contextual MHA and add a table only to the indicated path. At equal parameter count and cache, the value-only model has lower validation NLL (2.617 vs. 2.644), while the key-only model has higher NIAH (60.13\% vs. 57.46\%). Using both tables improves both metrics over either single-table variant at the cost of a second table per layer.

\textbf{TALA (learned table)} replaces the embedding projection that forms the base with one learned 864-dimensional lookup shared across all layers. It keeps the layer-specific base readers and 216-dimensional contextual MLA stage. With the same cache as TALA, the larger lookup lowers validation NLL from 2.654 to 2.647 and yields lower Commonsense and NIAH averages.

\textbf{MHA+DirectKV} and \textbf{MLA+DirectKV} use two independent full-width linear maps of $\mathbf e_j$ at each layer, adding their outputs to the contextual key and value at the same source position. Their contextual paths use dense MHA and a 216-dimensional MLA latent, respectively. These variants test direct embedding readouts in place of a factorized token path. Both store full content keys and values in the reported cache. MHA+DirectKV improves all three quality metrics over MHA with additional parameters and compute. MLA+DirectKV uses a larger cache than TALA and scores lower on all three metrics in this comparison.

\paragraph{Base factorization.}
\textbf{MHA+Base} adds TALA's 864-dimensional embedding-derived base to standard dense contextual MHA. Each layer applies its own key and value up-projections to the base, and the resulting keys and values use one attention softmax. This tests the shared base alongside a dense contextual path. At the same cache as MHA, adding the base lowers validation NLL from 2.672 to 2.648 and raises the Commonsense mean from 45.26\% to 47.20\%, while NIAH decreases from 58.97\% to 55.48\%.

\textbf{TALA (split Base, 1:1)} and \textbf{TALA (split Base, 1:2)} each project $\mathbf e_j$ into independent key and value latents shared across layers. Their respective key/value widths are 432/432 and 288/576. Each layer decodes the two bases separately and keeps a single 216-dimensional contextual stage for both paths. This comparison holds total base width, model size, training compute, and KV capacity fixed. Allocating more of the base to values raises NIAH from 65.48\% to 72.41\%, with similar validation NLL and a slightly lower Commonsense mean.

\paragraph{Context-augmented base writers.}
These four variants test whether the base benefits from local context before the first attention layer. All keep the base width at 864 and the contextual stage width at 216, with the same persistent cache of 10,656 bytes per token. Context is restricted to the current packed document.

\textbf{TALA (bigram)} adds a learned linear projection of the preceding embedding to the current-token base. \textbf{TALA (conv)} applies a width-four causal depthwise filter over the current and three preceding embeddings before the base projection. Out-of-document embeddings contribute zero in both variants. \textbf{TALA (gated Base)} averages the available previous three embeddings and projects that mean into an elementwise $1+\tanh(\cdot)$ gate on the current-token base. The context vector is zero at document starts. \textbf{TALA (hash)} maps the ordered previous/current token pair to a row of a learned $65{,}536\times64$ table. A learned projection maps the row to width 864 and adds it to the current-token base. A start-of-document sentinel supplies the previous-token ID at boundaries.

The additions test complementary ways to use local context, from injecting neighboring embeddings to modulating the token's features or learning a discrete pair representation. The gated base gives the highest NIAH in this group (77.14\% vs. 71.70\% for the token-only base), with the same displayed validation NLL. The hashed variant reaches the lowest NLL in the group (2.651) and has the lowest NIAH (62.89\%). All four preserve cache size, and their relative quality depends on the metric.

\Needspace{10\baselineskip}
\begingroup
\TALATableStyle
\setlength{\tabcolsep}{\TALAPreprintValue{3pt}{2pt}}
\begin{longtable}{@{}p{.24\linewidth}>{\raggedright\arraybackslash}p{.20\linewidth}rrrrrr@{}}
\caption{Non-loop Transformer variants trained on 10B tokens. KV is whole-model persistent BF16 bytes per token at 4K, and FLOPs are in $10^{18}$. CS and NIAH are the two mean benchmark scores (\%). The fixed-cache allocation sweep uses MLA and TALA as endpoints.}\label{tab:nonloop-all}\\
\toprule
Model & Configuration & \multicolumn{1}{c}{Params (M)} & \multicolumn{1}{c}{FLOPs (EF)} & \multicolumn{1}{c}{KV} & \multicolumn{1}{c}{\TALAMin{NLL}} & \multicolumn{1}{c}{\TALAMax{CS}} & \multicolumn{1}{c}{\TALAMax{NIAH}} \\
\midrule
\endfirsthead
\multicolumn{8}{@{}l}{\small\tablename~\thetable\ (continued)} \\
\toprule
Model & Configuration & \multicolumn{1}{c}{Params (M)} & \multicolumn{1}{c}{FLOPs (EF)} & \multicolumn{1}{c}{KV} & \multicolumn{1}{c}{\TALAMin{NLL}} & \multicolumn{1}{c}{\TALAMax{CS}} & \multicolumn{1}{c}{\TALAMax{NIAH}} \\
\midrule
\endhead
\bottomrule
\endfoot
\multicolumn{8}{@{}l}{\textit{Standard attention and cross-layer sharing}} \\*
MHA & Independent dense KV & 452.752 & 29.777 & 84,096 & 2.672 & 45.26 & 58.97 \\
GQA-3 & 3 KV heads & 412.939 & 27.388 & 14,976 & 2.699 & 44.81 & 42.94 \\
GQA-6 & 6 KV heads & 420.902 & 27.866 & 28,800 & 2.691 & 45.02 & 53.81 \\
MLA & Independent MLA, small cache & 421.404 & 27.896 & 10,656 & 2.699 & 45.21 & 56.94 \\
MLA (wide) & Independent MLA, large cache & 476.661 & 31.210 & 42,624 & 2.684 & 45.93 & 59.60 \\
CLA-2 & Shared KV over 2 layers & 428.864 & 28.344 & 42,624 & 2.683 & 45.24 & 43.18 \\
CLLA-2 & Shared latent over 2 layers & 432.353 & 28.553 & 10,656 & 2.678 & 45.47 & 58.80 \\
YOCO-SWA & Global--local memory & 418.912 & 27.224 & 14,976 & 2.681 & 45.81 & 21.79 \\
TALA & Shared base + stage (864/216) & 455.245 & 29.926 & 10,656 & 2.654 & 46.87 & 71.70 \\
TALA (realloc.) & Reallocated base/stage (944/224) & 458.379 & 30.114 & 10,656 & 2.650 & 47.01 & 70.71 \\
\midrule
\multicolumn{8}{@{}l}{\textit{Cache-compression tiers}} \\*
TALA (16$\times$) & Matched small cache & 427.318 & 28.251 & 5,376 & 2.673 & 46.03 & 71.38 \\
MLA (16$\times$) & Matched small cache & 412.278 & 27.348 & 5,376 & 2.709 & 44.45 & 60.65 \\
\shortstack[l]{MLA (4$\times$ cache,\\16$\times$ tier)} & 4-fold larger cache & 440.155 & 29.020 & 21,504 & 2.684 & 45.39 & 47.37 \\
TALA (32$\times$) & Matched small cache & 412.424 & 27.357 & 2,560 & 2.709 & 44.40 & 52.00 \\
MLA (32$\times$) & Matched small cache & 407.410 & 27.056 & 2,560 & 2.749 & 42.80 & 36.77 \\
\shortstack[l]{MLA (4$\times$ cache,\\32$\times$ tier)} & 4-fold larger cache & 420.685 & 27.852 & 10,240 & 2.694 & 45.28 & 54.06 \\
\midrule
\multicolumn{8}{@{}l}{\textit{Base/stage capacity allocation}} \\*
TALA (288/248) & Base/stage widths 288/248 & 432.684 & 28.572 & 10,656 & 2.666 & 45.97 & 62.52 \\
TALA (576/232) & Base/stage widths 576/232 & 443.965 & 29.249 & 10,656 & 2.657 & 46.55 & 76.97 \\
\midrule
\multicolumn{8}{@{}l}{\textit{Token tables and direct token readouts}} \\*
MemAtten & Separate learned token memory & 1,470.995 & 28.344 & 84,096 & 2.623 & 46.56 & 48.81 \\
MHA+KV tables & Independent K and V tables & 2,537.013 & 29.777 & 84,096 & 2.606 & 46.52 & 63.68 \\
MHA+VTable & Value table only & 1,494.882 & 29.777 & 84,096 & 2.617 & 46.52 & 57.46 \\
MHA+KTable & Key table only & 1,494.882 & 29.777 & 84,096 & 2.644 & 45.21 & 60.13 \\
TALA (learned table) & Learned base lookup & 497.672 & 29.866 & 10,656 & 2.647 & 46.38 & 62.31 \\
MHA+DirectKV & Direct token K/V readout & 500.528 & 32.643 & 84,096 & 2.663 & 45.94 & 60.61 \\
MLA+DirectKV & Direct token K/V readout & 466.193 & 30.583 & 84,096 & 2.677 & 46.01 & 61.65 \\
\midrule
\multicolumn{8}{@{}l}{\textit{Base factorization}} \\*
MHA+Base & Shared latent with dense MHA & 489.580 & 31.987 & 84,096 & 2.648 & 47.20 & 55.48 \\
TALA (split Base, 1:1) & Separate K/V base, 1:1 & 437.329 & 28.851 & 10,656 & 2.659 & 46.46 & 65.48 \\
TALA (split Base, 1:2) & Separate K/V base, 1:2 & 437.329 & 28.851 & 10,656 & 2.658 & 46.23 & 72.41 \\
\midrule
\multicolumn{8}{@{}l}{\textit{Context-augmented base writers}} \\*
TALA (bigram) & Adjacent token input & 456.241 & 29.986 & 10,656 & 2.655 & 46.39 & 73.80 \\
TALA (conv) & Causal depthwise convolution & 455.249 & 29.926 & 10,656 & 2.654 & 46.29 & 72.87 \\
TALA (gated Base) & Context-dependent gate & 456.241 & 29.986 & 10,656 & 2.654 & 46.66 & 77.14 \\
TALA (hash) & Hashed bigram table & 459.495 & 29.929 & 10,656 & 2.651 & 46.80 & 62.89 \\
\end{longtable}
\endgroup

\begingroup
\TALATableStyle
\setlength{\TALAScoreWidth}{28pt}
\begin{longtable}{@{}l*{10}{Q}@{}}
\caption{Nine language-benchmark scores (\%) for the 32 non-loop Transformer configurations trained on 10B tokens. Task abbreviations follow Table~\ref{tab:scaling-selected}.}\label{tab:nonloop-commonsense}\\
\toprule
\textbf{Model} & \TALAScoreHead{L-O} & \TALAScoreHead{L-S} & \TALAScoreHead{HS} & \TALAScoreHead{PIQA} & \TALAScoreHead{WG} & \TALAScoreHead{A-E} & \TALAScoreHead{A-C} & \TALAScoreHead{SciQ} & \TALAScoreHead{MMLU} & \TALAScoreHead[\uparrow]{Avg.} \\
\midrule
\endfirsthead
\multicolumn{11}{@{}l}{\small\tablename~\thetable\ (continued)} \\
\toprule
\textbf{Model} & \TALAScoreHead{L-O} & \TALAScoreHead{L-S} & \TALAScoreHead{HS} & \TALAScoreHead{PIQA} & \TALAScoreHead{WG} & \TALAScoreHead{A-E} & \TALAScoreHead{A-C} & \TALAScoreHead{SciQ} & \TALAScoreHead{MMLU} & \TALAScoreHead[\uparrow]{Avg.} \\
\midrule
\endhead
\bottomrule
\endfoot
TALA & 35.80 & 30.16 & 42.59 & 67.30 & 52.72 & 55.18 & 29.95 & 79.00 & 29.09 & 46.87 \\
MLA & 34.00 & 27.52 & 40.67 & 67.03 & 52.72 & 53.70 & 28.33 & 74.40 & 28.54 & 45.21 \\
MLA (wide) & 35.07 & 28.51 & 41.54 & 66.16 & 52.96 & 54.34 & 29.10 & 77.20 & 28.55 & 45.93 \\
MHA+DirectKV & 35.38 & 29.79 & 42.33 & 66.27 & 51.38 & 53.24 & 28.07 & 78.00 & 29.01 & 45.94 \\
MHA & 33.63 & 26.63 & 41.82 & 67.19 & 52.49 & 53.83 & 28.33 & 74.30 & 29.15 & 45.26 \\
MemAtten & 34.93 & 29.48 & 43.91 & 68.17 & 53.04 & 54.84 & 30.29 & 74.70 & 29.68 & 46.56 \\
MHA+KV tables & 35.77 & 29.42 & 44.36 & 68.72 & 50.91 & 54.34 & 29.18 & 76.60 & 29.42 & 46.52 \\
MHA+VTable & 35.86 & 29.28 & 43.56 & 67.41 & 53.20 & 55.05 & 29.01 & 75.80 & 29.48 & 46.52 \\
MHA+KTable & 33.42 & 26.53 & 42.73 & 67.63 & 51.14 & 53.75 & 27.90 & 75.20 & 28.55 & 45.21 \\
TALA (learned table) & 35.53 & 29.58 & 42.71 & 66.97 & 52.88 & 55.01 & 27.90 & 77.80 & 29.01 & 46.38 \\
MLA+DirectKV & 35.57 & 30.16 & 42.03 & 67.68 & 53.04 & 54.08 & 27.65 & 75.20 & 28.66 & 46.01 \\
TALA (16$\times$) & 36.29 & 27.67 & 42.38 & 67.19 & 53.20 & 53.24 & 29.01 & 76.30 & 28.97 & 46.03 \\
MLA (16$\times$) & 32.10 & 27.09 & 40.45 & 66.27 & 50.83 & 52.36 & 29.01 & 73.50 & 28.46 & 44.45 \\
\shortstack[l]{MLA (4$\times$ cache,\\16$\times$ tier)} & 34.39 & 26.92 & 41.31 & 65.83 & 53.83 & 53.41 & 29.52 & 74.90 & 28.44 & 45.39 \\
TALA (32$\times$) & 33.15 & 27.42 & 40.50 & 66.59 & 52.64 & 51.14 & 27.82 & 72.50 & 27.80 & 44.40 \\
MLA (32$\times$) & 30.35 & 22.41 & 39.06 & 66.32 & 48.93 & 52.86 & 27.65 & 69.50 & 28.08 & 42.80 \\
\shortstack[l]{MLA (4$\times$ cache,\\32$\times$ tier)} & 33.92 & 27.52 & 41.05 & 67.14 & 52.49 & 53.87 & 27.73 & 75.40 & 28.39 & 45.28 \\
MHA+Base & 37.30 & 32.18 & 43.06 & 66.81 & 53.20 & 54.21 & 28.92 & 80.20 & 28.91 & 47.20 \\
TALA (split Base, 1:1) & 35.16 & 30.12 & 42.16 & 67.25 & 52.09 & 54.17 & 29.18 & 79.20 & 28.83 & 46.46 \\
TALA (split Base, 1:2) & 35.46 & 28.66 & 42.30 & 67.19 & 52.25 & 54.34 & 29.44 & 77.60 & 28.83 & 46.23 \\
TALA (288/248) & 35.34 & 29.85 & 41.94 & 65.67 & 53.28 & 53.54 & 28.41 & 77.50 & 28.25 & 45.97 \\
TALA (576/232) & 37.22 & 31.11 & 42.76 & 67.52 & 51.30 & 54.00 & 28.84 & 77.20 & 29.03 & 46.55 \\
TALA (bigram) & 36.08 & 30.41 & 42.74 & 67.74 & 52.17 & 53.79 & 29.18 & 76.40 & 29.03 & 46.39 \\
TALA (conv) & 35.73 & 29.83 & 42.73 & 66.97 & 53.04 & 53.11 & 28.33 & 78.10 & 28.76 & 46.29 \\
TALA (gated Base) & 35.59 & 30.88 & 42.94 & 67.74 & 53.67 & 54.92 & 28.58 & 77.00 & 28.61 & 46.66 \\
TALA (hash) & 35.82 & 30.66 & 42.66 & 67.52 & 53.83 & 53.41 & 28.75 & 79.50 & 29.08 & 46.80 \\
GQA-3 & 32.91 & 26.04 & 40.69 & 66.27 & 52.80 & 52.57 & 27.99 & 75.60 & 28.46 & 44.81 \\
GQA-6 & 33.15 & 26.28 & 41.15 & 66.54 & 53.59 & 52.74 & 29.10 & 73.70 & 28.96 & 45.02 \\
CLA-2 & 33.09 & 25.42 & 41.52 & 67.79 & 51.14 & 53.41 & 29.86 & 76.20 & 28.71 & 45.24 \\
CLLA-2 & 34.54 & 27.48 & 41.86 & 66.27 & 51.38 & 54.67 & 29.44 & 75.10 & 28.50 & 45.47 \\*
YOCO-SWA & 33.67 & 26.49 & 42.12 & 67.03 & 54.14 & 54.80 & 29.52 & 75.90 & 28.64 & 45.81 \\*
TALA (realloc.) & 37.03 & 31.50 & 43.09 & 67.52 & 55.41 & 53.37 & 29.10 & 76.80 & 29.28 & 47.01 \\
\end{longtable}
\endgroup

\begingroup
\TALATableStyle
\setlength{\TALAScoreWidth}{28pt}
\begin{longtable}{@{}l*{7}{Q}@{}}
\caption{Six RULER-NIAH task scores (\%) at 4K for the 32 non-loop Transformer configurations trained on 10B tokens. Avg. is the unweighted task mean.}\label{tab:nonloop-ruler}\\
\toprule
\textbf{Model} & \TALAScoreHead{S1} & \TALAScoreHead{S2} & \TALAScoreHead{S3} & \TALAScoreHead{MK1} & \TALAScoreHead{MV} & \TALAScoreHead{MQ} & \TALAScoreHead[\uparrow]{Avg.} \\
\midrule
\endfirsthead
\multicolumn{8}{@{}l}{\small\tablename~\thetable\ (continued)} \\
\toprule
\textbf{Model} & \TALAScoreHead{S1} & \TALAScoreHead{S2} & \TALAScoreHead{S3} & \TALAScoreHead{MK1} & \TALAScoreHead{MV} & \TALAScoreHead{MQ} & \TALAScoreHead[\uparrow]{Avg.} \\
\midrule
\endhead
\bottomrule
\endfoot
TALA & 99.80 & 100.00 & 95.00 & 57.60 & 35.95 & 41.85 & 71.70 \\
MLA & 99.40 & 82.20 & 67.00 & 41.40 & 25.90 & 25.75 & 56.94 \\
MLA (wide) & 99.00 & 96.80 & 89.20 & 30.00 & 21.05 & 21.55 & 59.60 \\
MHA+DirectKV & 62.00 & 98.80 & 97.60 & 51.00 & 29.75 & 24.50 & 60.61 \\
MHA & 92.00 & 91.20 & 87.40 & 36.00 & 24.45 & 22.75 & 58.97 \\
MemAtten & 59.40 & 59.60 & 63.00 & 42.60 & 33.50 & 34.75 & 48.81 \\
MHA+KV tables & 90.00 & 97.60 & 93.00 & 49.80 & 27.50 & 24.15 & 63.68 \\
MHA+VTable & 54.40 & 98.40 & 88.20 & 47.00 & 33.70 & 23.05 & 57.46 \\
MHA+KTable & 77.00 & 94.20 & 85.80 & 49.40 & 28.20 & 26.15 & 60.13 \\
TALA (learned table) & 98.80 & 94.60 & 73.20 & 52.20 & 20.80 & 34.25 & 62.31 \\
MLA+DirectKV & 98.60 & 99.60 & 65.80 & 48.80 & 26.50 & 30.60 & 61.65 \\
TALA (16$\times$) & 98.60 & 100.00 & 91.00 & 51.60 & 41.95 & 45.15 & 71.38 \\
MLA (16$\times$) & 98.60 & 99.80 & 87.40 & 49.00 & 19.30 & 9.80 & 60.65 \\
\shortstack[l]{MLA (4$\times$ cache,\\16$\times$ tier)} & 96.20 & 76.80 & 68.60 & 34.60 & 4.15 & 3.85 & 47.37 \\
TALA (32$\times$) & 87.80 & 99.60 & 41.60 & 43.60 & 16.55 & 22.85 & 52.00 \\
MLA (32$\times$) & 54.60 & 83.40 & 29.40 & 29.80 & 11.90 & 11.50 & 36.77 \\
\shortstack[l]{MLA (4$\times$ cache,\\32$\times$ tier)} & 99.60 & 95.20 & 30.20 & 48.00 & 29.00 & 22.35 & 54.06 \\
MHA+Base & 86.40 & 90.80 & 26.40 & 53.40 & 37.70 & 38.20 & 55.48 \\
TALA (split Base, 1:1) & 98.80 & 100.00 & 55.80 & 48.20 & 47.85 & 42.20 & 65.48 \\
TALA (split Base, 1:2) & 90.40 & 100.00 & 96.20 & 54.80 & 49.10 & 43.95 & 72.41 \\
TALA (288/248) & 98.00 & 100.00 & 70.60 & 44.40 & 28.60 & 33.50 & 62.52 \\
TALA (576/232) & 99.80 & 100.00 & 93.40 & 64.20 & 51.85 & 52.55 & 76.97 \\
TALA (bigram) & 97.40 & 100.00 & 89.80 & 56.80 & 49.10 & 49.70 & 73.80 \\
TALA (conv) & 97.60 & 100.00 & 89.20 & 62.00 & 42.30 & 46.10 & 72.87 \\
TALA (gated Base) & 95.60 & 99.60 & 97.80 & 71.00 & 52.20 & 46.65 & 77.14 \\
TALA (hash) & 99.80 & 99.80 & 44.40 & 49.00 & 41.50 & 42.85 & 62.89 \\
GQA-3 & 98.80 & 89.40 & 2.80 & 27.80 & 19.05 & 19.80 & 42.94 \\
GQA-6 & 81.40 & 93.00 & 50.40 & 43.20 & 30.80 & 24.05 & 53.81 \\
CLA-2 & 98.40 & 77.00 & 3.40 & 48.40 & 23.55 & 8.35 & 43.18 \\
CLLA-2 & 96.80 & 99.60 & 62.80 & 36.20 & 29.80 & 27.60 & 58.80 \\*
YOCO-SWA & 24.80 & 29.00 & 20.40 & 25.00 & 17.50 & 14.05 & 21.79 \\*
TALA (realloc.) & 99.40 & 99.40 & 97.40 & 56.00 & 37.00 & 35.05 & 70.71 \\
\end{longtable}
\endgroup

\subsection{Training Trajectories}
\begin{figure}[!htbp]
\centering
\includegraphics[width=\linewidth]{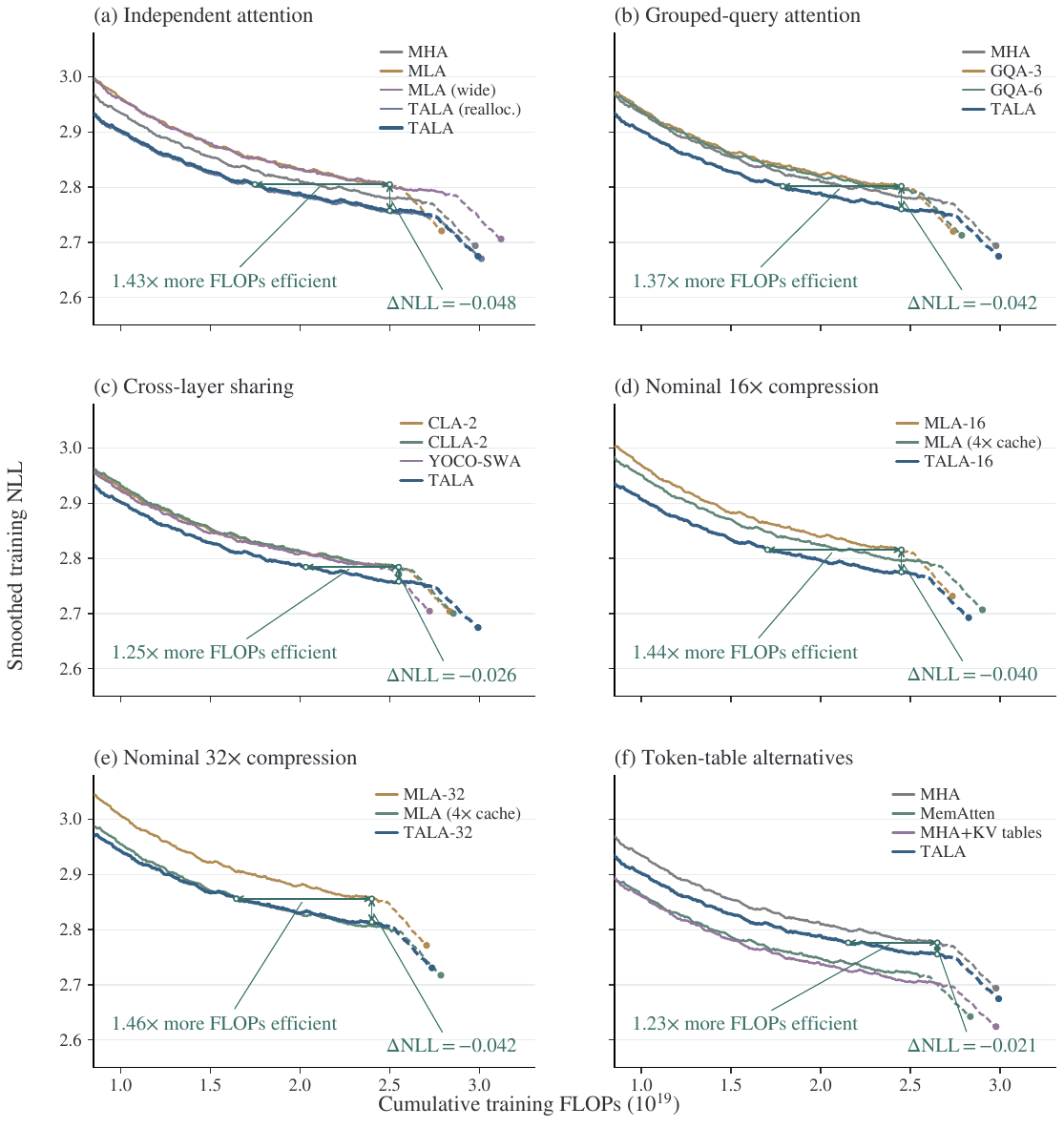}
\caption{Training trajectories of non-loop Transformers on 10B tokens. The annotated TALA comparisons use (a) MLA, (b) GQA-3, (c) CLLA-2, (d,e) cache-matched MLA, and (f) MHA as baselines. Smoothing and annotation conventions follow Figure~\ref{fig:fixed-depth-training-loss}. The wider MLA controls in (d,e) use four times the matched cache.}
\label{fig:nonloop-training-loss}
\end{figure}

Figure~\ref{fig:nonloop-training-loss} compares non-loop Transformers trained on 10B tokens. At matched smoothed training NLL in the stable phase, TALA improves FLOP efficiency by factors of 1.43, 1.37, and 1.25 relative to MLA, GQA-3, and CLLA-2, respectively. At nominal 16-fold and 32-fold compression, the corresponding factors relative to cache-matched MLA are 1.44 and 1.46. Panel (f) compares TALA with dense MHA alongside token-table alternatives.

\subsection{20B-token Pretraining}\label{app:nonloop_20b}
We extend MHA, MHA+DirectKV, MLA, and TALA to 20B tokens from their pre-decay checkpoints using the warmup--stable--decay schedule. TALA achieves the lowest validation NLL and highest downstream averages among the four architectures (Table~\ref{tab:nonloop-20b}). Tables~\ref{tab:nonloop-20b-commonsense} and~\ref{tab:nonloop-20b-ruler} give the individual task scores.
\begin{table}[htbp]
\centering
\TALATableStyle
\caption{Non-loop Transformers trained on 20B tokens. KV is whole-model persistent BF16 bytes per token at 4K. CS and NIAH are mean scores (\%). Bold and underlining mark the best and second-best quality metrics, respectively.}
\label{tab:nonloop-20b}
\begin{tabular}{@{}lrrrrrr@{}}
\toprule
& \multicolumn{3}{c}{Resources} & \multicolumn{3}{c}{Model quality} \\
\cmidrule(lr){2-4}\cmidrule(l){5-7}
\textbf{Model} & Params (M) & FLOPs (EF) & KV (B/token) & \TALAMin{NLL} & \TALAMax{CS} & \TALAMax{NIAH} \\
\midrule
MHA & 453 & 60 & 84,096 & 2.596 & 47.06 & 65.48 \\
MHA+DirectKV & 501 & 65 & 84,096 & \underline{2.586} & \underline{47.54} & \underline{67.38} \\
MLA & 421 & 56 & 10,656 & 2.618 & 46.85 & 60.82 \\
TALA & 455 & 60 & 10,656 & \textbf{2.582} & \textbf{48.19} & \textbf{75.81} \\
\bottomrule
\end{tabular}
\end{table}

\TALAFloat{table}{!htbp}{!htbp}
\centering
\TALATableStyle
\caption{Nine language-benchmark scores (\%) for the four non-loop Transformers trained on 20B tokens. Task abbreviations follow Table~\ref{tab:scaling-selected}.}\label{tab:nonloop-20b-commonsense}
\begin{tabular}{@{}lrrrrrrrrrr@{}}
\toprule
\textbf{Model} & \TALAMax{L-O} & \TALAMax{L-S} & \TALAMax{HS} & \TALAMax{PIQA} & \TALAMax{WG} & \TALAMax{A-E} & \TALAMax{A-C} & \TALAMax{SciQ} & \TALAMax{MMLU} & \TALAMax{Avg.} \\
\midrule
MHA & 35.28 & 27.87 & 44.95 & 68.39 & 52.49 & \textbf{56.52} & 30.46 & 77.40 & \textbf{30.14} & 47.06 \\
MHA+DirectKV & 37.38 & 30.60 & \textbf{46.09} & 68.23 & 52.88 & 55.47 & 29.52 & 77.90 & 29.80 & 47.54 \\
MLA & 36.04 & 29.69 & 44.33 & 67.57 & 52.17 & 55.89 & 29.52 & 77.20 & 29.27 & 46.85 \\
TALA & \textbf{37.78} & \textbf{31.94} & 45.99 & \textbf{69.10} & \textbf{53.51} & 56.36 & \textbf{30.55} & \textbf{78.90} & 29.60 & \textbf{48.19} \\
\bottomrule
\end{tabular}
\end{table}

\TALAFloat{table}{!htbp}{!htbp}
\centering
\TALATableStyle
\caption{Six RULER-NIAH task scores (\%) at 4K for the four non-loop Transformers trained on 20B tokens. Avg. is the unweighted task mean.}\label{tab:nonloop-20b-ruler}
\begin{tabular}{@{}lrrrrrrr@{}}
\toprule
\textbf{Model} & \TALAMax{S1} & \TALAMax{S2} & \TALAMax{S3} & \TALAMax{MK1} & \TALAMax{MV} & \TALAMax{MQ} & \TALAMax{Avg.} \\
\midrule
MHA & 97.40 & 98.00 & 93.20 & 44.40 & 31.75 & 28.10 & 65.48 \\
MHA+DirectKV & 97.60 & 99.00 & 87.80 & 47.80 & 40.60 & 31.50 & 67.38 \\
MLA & 99.40 & 99.00 & 89.40 & 45.60 & 15.15 & 16.35 & 60.82 \\
TALA & \textbf{99.60} & \textbf{100.00} & \textbf{94.00} & \textbf{63.00} & \textbf{53.20} & \textbf{45.05} & \textbf{75.81} \\
\bottomrule
\end{tabular}
\end{table}

\subsection{Evaluation-Time Attention Truncation in Frozen Non-Loop Transformers}\label{sec:tala-window-sensitivity}
The non-loop Transformers are pretrained on 10B tokens with full-history attention. With their weights frozen, we restrict per-layer causal attention at evaluation time to windows of 16, 64, 256, or 512 positions. Full-history evaluation provides the reference for each model. At window width $w$, query $i$ accesses $j\leq i$ with $i-j<w$ in the same document. All conditions use the same 430,611 validation targets with at least 512 positions of within-document history, across 492 documents.

At fixed KV capacity, a shared base reduces the penalty from restricting attention history (Figure~\ref{fig:tala-window-sensitivity}a,b). For a 16-position window, the NLL increase over full history is 2.16 for MLA and 0.84 for TALA. At 512 positions, the increases are approximately 0.14 and 0.057, respectively. TALA at 5.25 KiB per token also achieves lower subset NLL than MLA at 21 KiB per token across the tested windows (Figure~\ref{fig:tala-window-sensitivity}c).

\ifTALAPreprint
  \par\noindent\begin{minipage}{\linewidth}
  \captionsetup{hypcap=false,type=figure}
\else
  \begin{figure}[!ht]
\fi
  \centering
  \includegraphics[width=\TALAPreprintValue{.89}{.97}\linewidth]{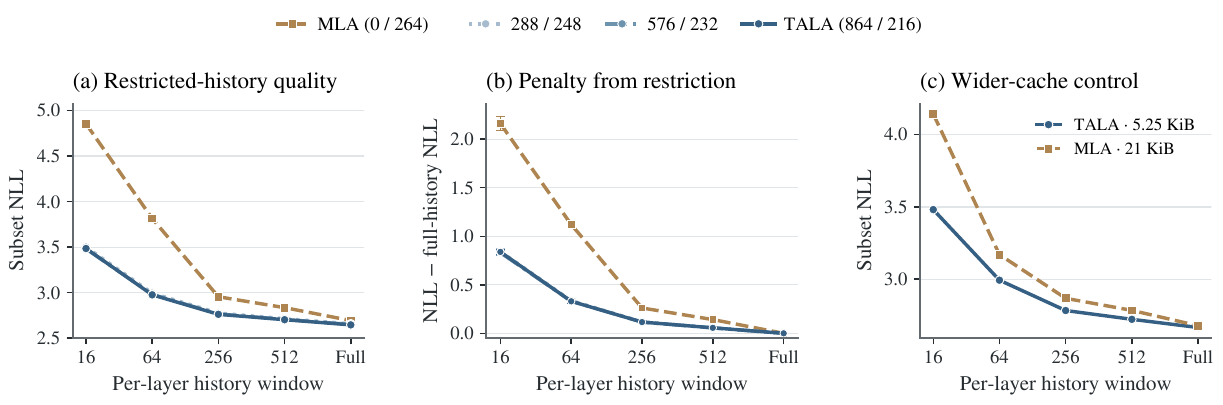}
  \caption{Evaluation-time truncation of per-layer attention history in frozen non-loop Transformers. (a) Validation NLL on the common long-history subset at fixed KV capacity. (b) NLL increase over full history, with paired 95\% document-bootstrap intervals (10,000 resamples). (c) TALA at 5.25 KiB/token versus MLA at 21 KiB/token.}
  \label{fig:tala-window-sensitivity}
\ifTALAPreprint
  \end{minipage}\par
\else
  \end{figure}
\fi

\section{Functional Specialization Analysis}\label{app:exp_mechanism_details}

\paragraph{Token-prototype protocol.}
We estimate token-conditional key prototypes on calibration documents and evaluate them on held-out documents with the realized queries (Section~\ref{sec:exp_mechanism}). We center content scores over legal source positions for each query, excluding positional scores. Frequency-matched token shuffling provides a null control. We report 95\% document-bootstrap intervals conditional on the frozen checkpoints. The looped analysis uses 20B equal-token models and 10,000 bootstrap replicates. In the non-loop analysis, the splits contain 215 calibration and 843 test documents. A minimum of 32 calibration occurrences yields 685 eligible token identities, and evaluation uses 1,955 fixed queries with up to 256 legal source positions per query. Each analysis has 492 contributing documents in its aggregate subset. Tables~\ref{tab:e72-full} and~\ref{tab:e67-full} report pooled predictability and absolute variances.

\TALAFloat{figure}{!htbp}{!tbp}
  \centering
  \includegraphics[width=0.97\linewidth]{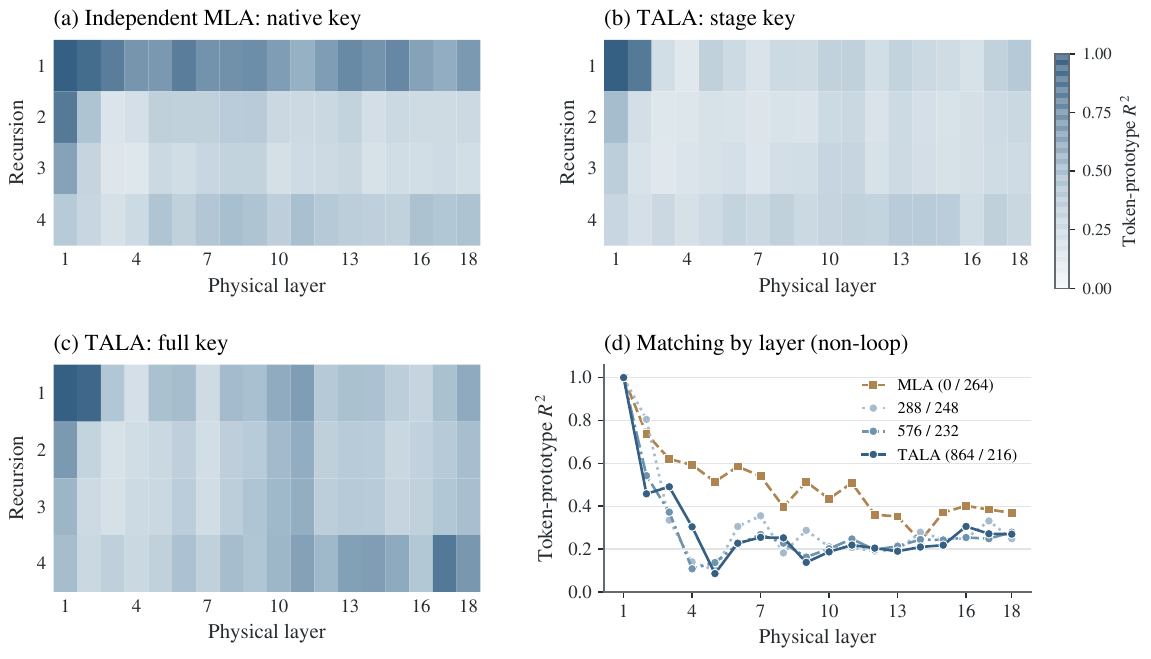}
  \caption{Layer-wise token-prototype $R^2$. (a--c) LT-MLA native keys and TALA stage and full keys across layers and recursions. (d) Non-loop stage-key predictability across layers at fixed KV capacity. Labels give base/stage widths.}
  \label{fig:tala-matching-detail}
\end{figure}

\begin{table}[htbp]
\centering
\TALATableStyle
\caption{Token-prototype analysis in looped Transformers. We report $R^2$ with 95\% document-bootstrap intervals and absolute score, prototype, and residual variances. Null $R^2$ uses frequency-matched token shuffling. Base-key $R^2$ equals one by construction.}
\label{tab:e72-full}
\begin{tabular}{lrrlrrrr}
\toprule
Model / component & R & $R^2$ & 95\% CI & Score var. & Proto. var. & Resid. var. & Null $R^2$\\
\midrule
TALA Base & 1 & 1.000 & [1.000, 1.000] & 0.197 & 0.197 & 0.000 & -1.012 \\
TALA Base & 2 & 1.000 & [1.000, 1.000] & 0.202 & 0.202 & 0.000 & -0.927 \\
TALA Base & 3 & 1.000 & [1.000, 1.000] & 0.209 & 0.209 & 0.000 & -0.906 \\
TALA Base & 4 & 1.000 & [1.000, 1.000] & 0.219 & 0.219 & 0.000 & -1.063 \\
TALA Stage & 1 & 0.280 & [0.276, 0.283] & 0.832 & 0.246 & 0.599 & -0.258 \\
TALA Stage & 2 & 0.212 & [0.208, 0.215] & 1.389 & 0.325 & 1.095 & -0.182 \\
TALA Stage & 3 & 0.210 & [0.207, 0.214] & 1.522 & 0.350 & 1.201 & -0.174 \\
TALA Stage & 4 & 0.278 & [0.273, 0.282] & 1.041 & 0.330 & 0.752 & -0.286 \\
LT-MLA native & 1 & 0.758 & [0.755, 0.760] & 0.779 & 0.596 & 0.189 & -0.665 \\
LT-MLA native & 2 & 0.314 & [0.311, 0.316] & 1.109 & 0.361 & 0.761 & -0.277 \\
LT-MLA native & 3 & 0.255 & [0.252, 0.257] & 1.264 & 0.335 & 0.943 & -0.238 \\
LT-MLA native & 4 & 0.386 & [0.382, 0.391] & 1.151 & 0.463 & 0.707 & -0.360 \\
\bottomrule
\end{tabular}

\end{table}

\begin{table}[htbp]
\centering
\TALATableStyle
\caption{Token-prototype analysis for the fixed-cache non-loop allocation sweep. We report absolute score, prototype, and residual variances and 95\% document-bootstrap intervals for $R^2$. Null $R^2$ uses frequency-matched token shuffling.}
\label{tab:e67-full}
\begin{tabular}{@{}rrrlrrrr@{}}
\toprule
Base & Stage & $R^2$ & 95\% CI & Score var. & Proto. var. & Resid. var. & Null $R^2$ \\
\midrule
0 & 264 & 0.423 & [0.419, 0.427] & 0.811 & 0.352 & 0.468 & -0.403 \\
288 & 248 & 0.216 & [0.212, 0.219] & 0.809 & 0.195 & 0.634 & -0.235 \\
576 & 232 & 0.192 & [0.189, 0.195] & 0.804 & 0.175 & 0.649 & -0.204 \\
864 & 216 & 0.190 & [0.187, 0.193] & 0.743 & 0.162 & 0.602 & -0.211 \\
\bottomrule
\end{tabular}
\end{table}

\paragraph{Layer-wise matching.}
Figure~\ref{fig:tala-matching-detail} shows matching across physical layers and recursions. Layer one of recursion one receives the initial token representation. Subsequent sites incorporate contextual computation. TALA stage keys have lower token-prototype $R^2$ than LT-MLA native keys at 60 of 72 layer--recursion sites, including 43 of 54 in recursions two through four. Across the four recursions, pooled full-key $R^2$ estimates and 95\% intervals are 0.460 [0.457, 0.464], 0.353 [0.350, 0.357], 0.352 [0.348, 0.355], and 0.446 [0.441, 0.450]. In the non-loop sweep, pooled stage-key predictability decreases with increasing base width.

\TALAFloat{figure}{!htbp}{!tbp}
  \centering
  \includegraphics[width=0.97\linewidth]{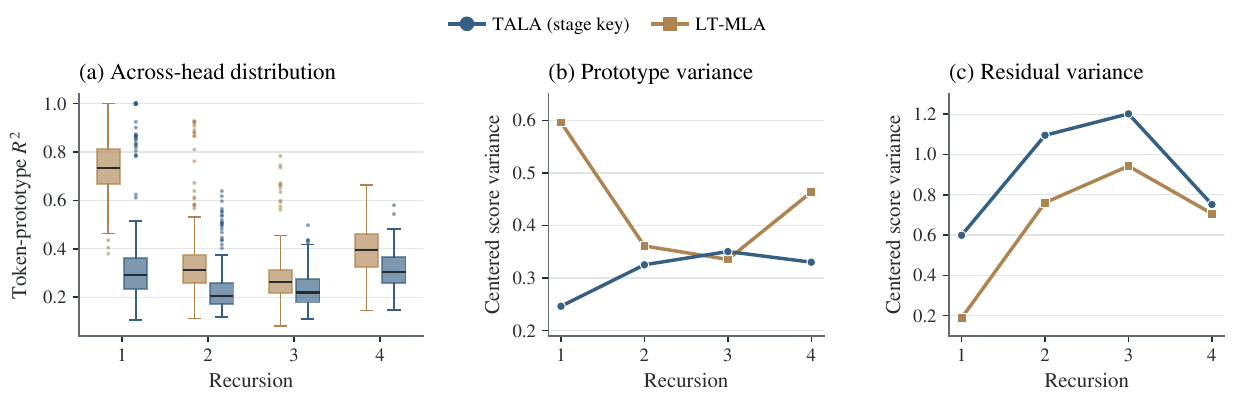}
  \caption{Head-level variation and absolute score variances in looped Transformers. (a) Token-prototype $R^2$ across 324 layer--head sites per component and recursion. Whiskers extend to 1.5 interquartile ranges. (b,c) Observation-weighted prototype and residual score variances, respectively.}
  \label{fig:tala-head-distribution}
\end{figure}

\paragraph{Head-wise variation and absolute variance.}
Figure~\ref{fig:tala-head-distribution}a shows the unweighted distribution across 324 layer--head sites per component and recursion. Panels (b,c) report observation-weighted pooled prototype and residual variances. The pooled stage-key $R^2$ is lower than the native-key $R^2$ in all four recursions, while stage prototype variance is slightly higher in recursion three. In the non-loop sweep, prototype variance decreases as the shared base grows (Table~\ref{tab:e67-full}). Total score variance includes both component variances and twice their covariance.

\end{document}